\documentclass[review]{elsarticle}

\usepackage{lineno}
\usepackage{hyperref}
\usepackage{color}
\usepackage{amssymb, amsmath}
\usepackage{amsfonts, mathrsfs, marvosym, dsfont}
\usepackage{bbold} 
\usepackage[linesnumbered,lined,boxed,commentsnumbered]{algorithm2e}
\RestyleAlgo{ruled} %
\modulolinenumbers[5]
\usepackage{hyperref}
\usepackage{makecell}
\usepackage{subcaption}
\usepackage{multirow}
\usepackage{tablefootnote}

\usepackage{todonotes}
\usepackage{url}

\newcommand{\blueCN}[1]{\textcolor[rgb]{0.00,0.00,0.00}{#1}}
\newcommand{\blueC}[1]{\textcolor[rgb]{0.00,0.00,0.00}{#1}}

\newcommand{\greenC}[1]{\textcolor[rgb]{0,0,0}{#1}}

\begin{document}

\newcommand{\argmax}[1] {\underset{#1}{\operatorname{argmax}}}
\newcommand{\argsup}[1] {\underset{#1}{\operatorname{argsup}}}
\newcommand{\argmin}[1] {\underset{#1}{\operatorname{argmin}}}
\newcommand{\arginf}[1] {\underset{#1}{\operatorname{arginf}}}

\newcommand{\trans}[1] {\ensuremath{\left( #1 \right)^\intercal}}
\newcommand{\inv}[1]   {\ensuremath{\left( #1 \right)^{-1}}}
\newcommand{\Space}[1] {\ensuremath{\mathcal{S}\left( #1 \right)}}

\newcommand{\chose}[1]    {\textcolor{green}{#1}}
\newcommand{\tomodify}[1] {\textcolor{blue}{#1}}

\newcommand{\pdf}[1]  {p\left(#1\right)}
\newcommand{\pdfc}[2]  {\ensuremath{p\left( #1 \left| #2 \right.\right)}}

\newcommand{\set}[1]    {\left\{#1\right\}}
\newcommand{\seq}[1]    {\left(#1\right)}
\newcommand{\interv}[1] {\left[#1\right]}

\newcommand{\bs}[1]    {\ensuremath{\boldsymbol{#1}}}

\newcommand{\Norm}       {\mathcal{N}}
\newcommand{\FollowNorm} {\thicksim\Norm}
\newcommand{\Ln}         {\mathrm{ln}}
\newcommand{\Card}[1]    {Card\left(#1\right)}
\newcommand{\Real}       {\mathds{R}}

\newcommand{\x}       {\bs{x}}
\newcommand{\X}       {\bs{X}}
\newcommand{\xn}      {\x_{n}}
\newcommand{\xun}     {\x_{1}}
\newcommand{\xdeux}   {\x_{2}}
\newcommand{\xN}      {\x_{N}}
\newcommand{\xunN}    {\x_{1}^N}

\newcommand{\Y}       {\bs{Y}}
\newcommand{\y}       {\bs{y}}

\newcommand{\U}       {\bs{U}}
\renewcommand{\u}     {\bs{u}}

\newcommand{\z}       {\mathrm{z}}
\newcommand{\Z}       {\mathrm{Z}}
\newcommand{\zn}      {\z_{n}}
\newcommand{\zun}     {\z_{1}}
\newcommand{\zdeux}   {\z_{2}}
\newcommand{\zN}      {\z_{N}}
\newcommand{\zunN}    {\z_{1}^N}

\newcommand{\bsz}       {\mathrm{\bs{z}}}

\newcommand{\bsmu}    {\bs{\mu}}
\newcommand{\bsSigma} {\bs{\Sigma}}

\newcommand{\bsTheta} {\bs{\Theta}}
\newcommand{\bstheta} {\bs{\theta}}
\newcommand{\hattheta}{\hat{\theta}}

\newcommand{\bspi} {\bs{\pi}}

\renewcommand{\H} {\bs{H}}
\newcommand{\G}   {\bs{G}}
\newcommand{\F}   {\bs{F}}

\newcommand{\Gammafun}[1]      {\Gamma\left(#1\right)}
\newcommand{\incGammafun}[2]   {\large{\gamma}\left(#1,#2\right)}
\newcommand{\Betafun}[2]       {B\left(#1,#2\right)}
\newcommand{\incBetafun}[3]    {I_{#1}\left(#2,#3\right)}
\newcommand{\errorfun}[1]      {erf\left(#1\right)}
\renewcommand{\exp}[1]         {exp\left(#1\right)}
\newcommand{\hypergeofun}[4]   {{_2\mathcal{F}_1}\left(#1,#2,#3,#4\right)}

\begin{frontmatter}

\title{Copula-based mixture model identification for subgroup clustering \blueC{with imaging applications}}

\author[creatis]{Fei Zheng}

\author[creatis,iuf]{Nicolas Duchateau\corref{mycorrespondingauthor}}
\cortext[mycorrespondingauthor]{Corresponding author}

\address[creatis]{Univ Lyon, INSA‐Lyon, Université Claude Bernard Lyon 1, UJM-Saint Etienne, CNRS, Inserm, CREATIS UMR 5220, U1294, F‐69621, LYON, France \vspace{2mm}}

\address[iuf]{Institut Universitaire de France (IUF)}

\begin{abstract}
Model-based clustering techniques have been widely applied to various application areas, while most studies focus on canonical mixtures with unique component distribution form. However, this strict assumption is often hard to satisfy. In this paper, we consider the more flexible Copula-Based Mixture Models (CBMMs) for clustering, which allow heterogeneous component distributions composed by flexible choices of marginal and copula forms. \blueC{More specifically, we propose an adaptation of the Generalized Iterative Conditional Estimation (GICE) algorithm to identify the CBMMs in an unsupervised manner, where the marginal and copula forms and their parameters are estimated iteratively.}

GICE is adapted from its original version developed for switching Markov model identification with the choice of realization time. Our CBMM-GICE clustering method is then tested on synthetic \blueC{two-cluster} data \blueCN{(N=2000 samples)} with discussion of the factors impacting its convergence. Finally, \blueC{it is compared to the Expectation Maximization identified mixture models with unique component form \blueCN{on the entire} MNIST database \greenC{(N=70000)}, and on real cardiac \blueCN{magnetic resonance} data \blueCN{(N=276)} to illustrate its \blueCN{value for imaging applications}.}
\end{abstract}


\end{frontmatter}

\section{Introduction}
\label{sec:Introduction}

Finite mixture models are very relevant tools to model the distribution of data samples in a population, and in particular for unsupervised learning problems such as clustering, outlier detection, or even generating new samples from a given population \cite{mclachlan2019finite}. 
The most widely applied mixture models are Gaussian Mixture Models (GMMs), which
assume that each subgroup of the population follows a specific multivariate Gaussian distribution. GMMs can be identified through the Expectation-Maximization (EM) algorithm which iteratively finds the local maximum data likelihood \cite{mclachlan2007algorithm}.
GMMs have been commonly chosen mainly because they offer a closed-form solution for the maximization step in the EM algorithm, and better interpretability thanks to its easy marginalization which allows visualizing the model in a given dimension. 
However, GMMs are obviously not a good choice for fitting a population composed of non-elliptical clusters as often found in medical \blueCN{applications \cite{Banfield_1993}}, since they may overestimate the number of components to approximate the real distribution \blueCN{\cite{Baudry_2010}}. 
Mixtures of alternative component densities have been explored during last decades for specific non-Gaussian application situations. For example, Weibull mixtures for learning the distribution of the cotton fiber length \cite{kuang2015generating}, Gamma or Beta mixtures for SAR image modeling \cite{ma2011bayesian, liu2019bayesian}, Dirichlet mixtures for skin color modeling \cite{bouguila2004unsupervised}, and Student's t mixtures for image segmentation \cite{sfikas2007robust}. There have also been attempts for more flexible component distribution by introducing skewness \cite{lee2014finite}, 
or multidimensional scale variables \cite{forbes2014new}. 

In spite of the added flexibility offered by the above mixture models, they assume that i) all the components follow the same family of distribution, ii) all dimensions of subgroups follow the same family of marginal distribution. Some heterogeneous components mixtures have been proposed to overcome the first constraint \cite{li2016statistical, huang2019probability}. However, the second assumption may also not be satisfied in practice when performing multivariate analysis. For example in medical analytics, the blood pressure usually follows a normal or skewed normal distribution while the medical cost usually follows a log-normal distribution or a mixture of log-normal and normal distributions \cite{fujimaki2011online}. Similarly, as in one of the application examples used in this paper, myocardial infarct shapes across subjects and image slices do not necessarily follow a normal distribution, as these are constrained by the anatomy and the physiology of ischemia-reperfusion mechanisms \cite{Duchateau:FrontCV:20233}.
To address this issue, ``copulas'' have been introduced into the mixture model family \cite{nelsen2007introduction}. 
According to Sklar's theorem, any multivariate joint distribution can be decomposed into its marginal distributions and a copula which represents the dependency structure among variables from their marginal distributions \cite{sklar1959fonctions}. 
The resulting copula-based mixture models (CBMMs), alternatively named ``Mixture models with Heterogeneous Marginals and Copulas (MHMC)'' \cite{fujimaki2011online}, allow us to model both heterogeneous types of marginal distributions and heterogeneous dependencies among variables separately, thus offers more flexibility and better approximation to the density of data in practice. The name ``copula-based mixture models'' is preferred here to stay brief as no confusion will be introduced.

Although the copula-based mixtures are very flexible for density estimation, \blueC{they} are not much explored so far for applications due to the difficulty of selecting marginal distributions and copulas.
Most of the work determines the forms of \blueCN{marginals} and copulas intuitively or by data analysis such as a histogram or quantile-quantile plot, then estimate the model parameters through an EM-like iterative algorithm \cite{roy2014pair}. 
A classical way to deal with this model selection problem is to fit all combinations of possible marginal and copula forms \blueC{from a finite set of models}, then select the best-fit combination according to some criterion such as Bayesian Information Criterion (BIC). 
This two-step decision becomes computational-heavy when facing high-dimensional data given many candidate forms.  
\blueC{According to \cite{kosmidis2016model}, one needs to repeat the parameter estimation for $\binom{d + K -1}{K}$ combinations to finally choose the best-fit CBMM of $K$ components from a dictionary of $d = L^D M$ multivariate distributions forms,
for $D$-dimensional data,
with a candidate list of $L$ marginal forms and $M$ copula forms.} 

\subsubsection*{\blueC{Proposed approach and contributions}}

In this study, we propose an iterative algorithm for the identification of CBMMs \blueC{that overcomes these limitations. It estimates} optimized marginal and copula forms from a list of candidates as well as their parameters, by updating both the forms and parameters at each iteration. The proposed algorithm is a Generalized ICE (GICE) 
\cite{pieczynski2007convergence}
adapted for CBMMs inspired by the GICE algorithm for switching hidden Markov model identification \cite{zheng2020semi, derrode2016unsupervised}.
\blueC{It represents an original strategy for CBMM identification, which we thoroughly evaluate on synthetic and real data, including \blueCN{large populations from synthetic toy configurations and a popular public dataset, }and a real clinical clustering problem in medical imaging \blueCN{(infarct shape analysis from magnetic resonance images)}.}


\greenC{Similar works that also considers updating the form until the convergence of the algorithm for CBMM identification include the ``Online EMDL''\cite{fujimaki2011online} and ``VCMM''\cite{sahin2022vine}.  
Online EMDL estimates not only the marginal and copula forms but also the component number in an online manner. Even so, only three candidate marginal forms from exponential family and Gaussian copulas are explored in their study, which still limits the flexibility of the resulting mixture. VCMM is similar to GICE, while having the advantage that it can be applied on higher-dimensional data by incorporating a vine copula structure. VCMM estimates the marginal parameters at the beginning and the end of the algorithm instead of updating them continuously throughout the iterations as in GICE.}

\greenC{It is worth mentioning that recent studies have explored other flexible distribution models, such as neural copulas\cite{li2023nn} and Spline Quasi-interpolation-based methods\cite{tamborrino2024empirical}. In this article, we prefer to focus on the estimation of the conventional CBMM due to its interpretability in medical applications.}


\blueC{Our} paper is structured as follows. Section \ref{sec:Copula-based mixture models} presents the CBMM and specifies the identification problem.
Section \ref{sec:Iterative conditional estimation for model identification with form selection} describes the proposed iterative conditional estimation for CBMM identification, discusses the details of its configuration and convergence.
In Section \ref{sec:Experiments}, the proposed algorithm is tested on synthetic data for different scenarios considering different method configurations \blueC{(2 clusters)}, \blueC{\blueCN{on the entire} MNIST database (10 clusters),} and then applied to real cardiac image clustering. 
Finally, Section \ref{sec:conclusion} outlines the conclusions and proposals for further research.
\section{Copula-based mixture models}
\label{sec:Copula-based mixture models}

Let $\X = \seq{X^1,\cdots,X^D}$ be a $D$-dimensional random variable, the general mixtures model the Probability Density Function (PDF) of $\X$ as follows:
\begin{equation}
    \label{eq:general mixture}
    \pdf{\x} = \sum_{k=1}^K  \pdf{z=k} \pdfc{\x}{z=k},
\end{equation}
\sloppy where $\x$ denotes the realisation of $\X$, and $z$ is the realisation of $Z$, a categorical random variable
that takes its value in $\set{1,\cdots,K}$ indicating to which subgroup $\X$ belongs.
More concisely the model is often written as:
\begin{equation}
    \label{eq:concise general mixture}
    \pdf{\x} = \sum_{k=1}^K  \pi_k p_k\seq{\x},
\end{equation}
where $\pi_k = \pdf{z=k}$ is called component coefficient, \blueC{and} $p_k\seq{\x} = \pdfc{\x}{z=k}$. When all components follow a Gaussian distribution, namely $p_k\seq{\x}=\Norm\seq{\blueC \x | \bsmu_k, \bsSigma_k}$, 
the model becomes a classical GMM. 

\paragraph{Copulas} Let $\Y = \seq{Y^1, \cdots, Y^D}$ be a random vector valued in $\Real^D$, and $\y=\seq{y^1,\cdots,y^D}$ be a realisation of $\Y$. $F\seq{\y} = P\interv{Y^1\leq y^1, \cdots, Y^{\blueC D}\leq y^{\blueC D}}$ denotes the Cumulative \blueC{Distribution} Function (CDF) of $\Y$, and $F_1\seq{y^1}, \cdots, F_D\seq{y^D}$ denotes the univariate CDFs of $Y^1,\cdots,Y^D$ respectively. A copula is a CDF defined on $\interv{0, 1}^D$ such that marginal CDFs are uniform. According to Sklar's theorem \cite{nelsen2007introduction}, any multivariate distribution can be represented via its marginal distributions and a unique copula $C$ which links them:
\begin{equation}
    \label{eq:copula definition}
    F\seq{\y} = C\seq{F_1\seq{y^1},\cdots,F_D\seq{y^D}}.
\end{equation}
Assuming differentiable $F$ and $C$, and setting:
\begin{equation}
    \label{eq:copula pdf}
    c\seq{\u} = \frac{\partial^D}{\blueC{\partial u^1 \cdots \partial u^D}}C\seq{\u},
\end{equation}
where $\u = \seq{u^1,\cdots, u^D}$, taking the derivative of \blueC{\eqref{eq:copula definition}}, the PDF of $\Y$ can be represented with marginal densities $f_d$, $d\in\set{1,\cdots,D}$ and a copula density function $c$: 
\begin{equation}
    \label{eq:pdf with copula}
    f\seq{\y} = c\interv{F_1\seq{y^1},\cdots,F_D\seq{y^D}}\prod_{d=1}^D f_d\seq{y^d}.
\end{equation}

Applying \eqref{eq:pdf with copula} to decompose the component density $p_k\seq{\x}$ in the general mixture model \eqref{eq:concise general mixture}, we obtain the copula-based mixture model (CBMM):
\begin{equation}
    \label{eq:CBMMs}
    \pdf{\x} = \sum_{k=1}^K  \pi_k c_k\interv{F_{k,1}\seq{x^1},\cdots,F_{k,D}\seq{x^D}}\prod_{d=1}^D f_{k,d}\seq{x^d},
\end{equation}
Therefore, the choice of 
distribution of the $k$-th component
is ``decomposed'' 
into
choices of its marginal distributions $f_{k,d}$, and the choice of its copula $c_k$, $d\in\set{1,\cdots,D}$, $k\in\set{1,\cdots,K}$. Choosing all $f_{k,d}$ and $c_k$ to be Gaussian 
amounts to choose
Gaussian $p_k\seq{\x}$ and finds back a GMM. 

Let's now
consider the parameters in the model \eqref{eq:CBMMs}:
\begin{equation}
    \label{eq:CBMMs with params}
    \pdf{\x;\bsTheta} = \sum_{k=1}^K  \pi_k c_k\interv{F_{k,1}\seq{x^1;\theta_{k,1}},\cdots,F_{k,D}\seq{x^D;\theta_{k,D}}; \alpha_k} \prod_{d=1}^D f_{k,d}\seq{x^d;\theta_{k,d}},
\end{equation}
where 
$\theta_{k,d}$ represents the parameters of the marginal distribution 
for the $k$-th component in the $d$-th dimension, 
and 
$\alpha_k$ denotes the parameters of the copula 
for the $k$-th 
component. We can thus summarize the parameters as $\bsTheta = \set{\pi_k, \alpha_k, \bstheta_k}_{k=1}^K$, with component marginal parameter set $\bstheta_k=\set{\theta_{k,d}}_{d=1}^{D}$.
The choices of the marginal and copula forms lead to great flexibility of CBMMs and also bring the difficulty of the model identification. 
In practice, apart from the parameter estimation, we need to make appropriate decision among a bunch of candidate marginal and copula forms.

\section{Method: Generalized Iterative Conditional Estimation (GICE) for model identification with form selection}
\label{sec:Iterative conditional estimation for model identification with form selection}

Let $\x_1^N = \seq{\x_1, \cdots, \x_N}$ be the samples from a CBMM \eqref{eq:CBMMs with params}. Their subgroup labels $z_1^N=\seq{z_1,\cdots,z_N}$ are hidden. Given the component number $K$, the model identification means 
solving the following problems:
\begin{enumerate}
    \item For each \blueCN{marginal} $f_{k,d}$, $d\in\set{1,\cdots,D}$, $k\in\set{1,\cdots,K}$, find its best-fit form $H_{k,d}$, and estimate its associate parameters $\theta_{k,d}$ from a list of candidate forms $\H = \set{H_1,\cdots,H_L}$. Each form $H_l$, $l\in\set{1,\cdots,L}$ is a parametric set of distributions $H_l=\set{f_{\theta\seq{l}}}_{\blueC{l\in\set{1,\cdots,L}}}$.
    \item For each copula $c_k$, $k\in\set{1,\cdots,K}$, find its best-fit form $G_{k}$, and estimate its associate parameters $\alpha_k$ from a list of candidate forms $\G = \set{G_1,\cdots,G_M}$. Each form $G_m$, $m\in\set{1,\cdots,M}$ is a parametric set of copulas $G_m=\set{c_{\alpha\seq{m}}}_{\blueC{m\in\set{1,\cdots,M}}}$.
    \item Estimate component coefficients $\bspi=\set{\pi_k}_{k=1}^K$ related to the hidden cluster labels $z_1^N$.
\end{enumerate}
We tackle the identification problem of the CBMMs through a Generalised ICE (GICE) framework with copula selection.  
ICE is an iterative method for parameter estimation in case of incomplete data, under two gentle assumptions: (i) there exist estimators for the parameters from the complete data ($\x_1^N, z_1^N$), and (ii) the ability to generate samples of $Z$ from $\pdfc{z}{\x}$ \cite{pieczynski2007convergence}. 
The ICE resembles \blueC{the} EM algorithm, but it does not aim to maximize the likelihood. It is generally easier to implement compared to EM, especially when facing complex \blueC{models. Besides, the} ``maximization'' step of EM often meets difficulties. 

To apply the ICE framework to our problem, we accept the following assumptions:
\begin{enumerate}
    \item There exists an estimator $\hat{\theta}(l)$ for estimating the parameters of the candidate marginal form $H_l$ , $\forall l\in\set{1,\cdots,L}$ from univariate data samples, and an estimator $\hat{\alpha}(m)$ for estimating the parameters of the
    candidate copula form $G_m$, $\forall m\in\set{1,\cdots,M}$ from multivariate data samples.
    \item There exist two decision rules, $\blueC{\Delta}^1$ deciding the best-fit marginal form from $\H$ which fits the given univariate samples $\seq{y_1,\cdots,y_Q}$, and $\blueC{\Delta}^2$ deciding the best-fit copula from $\G$ which fits the given multivariate samples $\seq{\y_1,\cdots,\y_Q}$. $Q$ denotes the number of samples.
\end{enumerate}

The first assumption is required by the original ICE, and the second assumption is added to enable the form selection of the algorithm. A two-step parameter estimation is considered in the first assumption to separate the marginal and the copula estimators. When estimating the parameters of a joint distribution, we first \blueC{need to define the number of clusters, then} estimate the marginal parameters, \blueC{and finally} estimate the copula parameters among estimated marginal CDFs. In consequence, the form selection of \blueCN{marginals} and copula can also be performed separately, which avoids the form combination issue involved when using simultaneous estimators.

\blueC{The proposed GICE algorithm starts from an initial guess of cluster distributions, then iteratively updates the distribution forms as well as the parameters based on grouped data with simulated cluster labels.}
More precisely, GICE starts with an initialization of parameters $\bsTheta^{(i=0)}=\set{\pi_k^{(0)}, \alpha_k^{(0)}, \bstheta_k^{(0)}}_{k=1}^K$, then iteratively runs the following steps: 
\begin{enumerate}
    \item \blueC{At iteration $i$, c}ompute the posteriors of the hidden variable $\pdfc{\zn}{\xn; \bsTheta^{(i-1)}}$, for $n=1,\cdots,N$ by:
    \begin{equation}
        \label{eq:GICE posterior of z}
        \pdfc{\zn=k}{\xn; \bsTheta^{(i-1)}} = \frac{\pi_k^{(i-1)} \pdfc{\xn}{\zn; \alpha_k^{(i-1)}, \bstheta_k^{(i-1)}} }{\sum_{k=1}^K \pi_k^{(i-1)} \pdfc{\xn}{\zn; \alpha_k^{(i-1)}, \bstheta_k^{(i-1)}}},
    \end{equation}
    then simulate $T$ times the hidden labels $\zunN$ \blueC{according to \eqref{eq:GICE posterior of z}}, name all the realizations $\hat{\bsz} = \seq{\hat{\bsz}^1,\cdots,\hat{\bsz}^T}$, with each realization set $\hat{\bsz}^t = \seq{\hat{z}_1^t,\cdots,\hat{z}_N^t}$, $t\in\set{1,\cdots,T}$.
    \item For each component $k$, \blueC{create the subgroup of samples by gathering the samples with their corresponding labels $\hat{\bsz}=k$. Then, for each created subgroup, select its marginal and copula forms from candidate lists $\H$, $\G$, and estimate their corresponding parameters $\bstheta_k$, $\alpha_k$ by executing:}
    \begin{enumerate}
        \item For all $t\in\set{1,\cdots,T}$, group data $\seq{\x_1^N}_{\hat{\z}^t=k}$ by taking the subset of $\x_1^N$ formed with $\x_n$ whose cluster label $\hat{z}_n^t=k$, then fuse the $T$     
        subsets corresponding to the same $k$ to obtain the entire subgroup samples:
        \begin{equation}
            \label{eq:GICE subgroup samples}
             \seq{\x_1^N}_{\hat{\bsz}=k} = \seq{\seq{\x_1^N}_{\hat{\z}^1=k},\cdots,\seq{\x_1^N}_{\hat{\z}^T=k}}.
        \end{equation}
        \item Estimate the coefficient of the $k$-th component by:
            \begin{equation}
            \label{eq:GICE update pi_k}
                \hat{\pi}_k = \blueCN{\frac{1}{NT}}\sum_{t=1}^T\sum_{n=1}^N\Card{\hat{z}_n^t=k},
            \end{equation}
        where $Card$ computes the cardinality of a given set.
        \item For all $d\in\set{1,\cdots,D}$, and all $l\in\set{1,\cdots,L}$, fit the marginal form $H_l$ to $\seq{x_1^N}_{\hat{\bsz}=k}^d$
        \greenC{(namely, the $d$-th dimension of $\seq{x_1^N}_{\hat{\bsz}=k}$)} by predefined marginal estimator, and get candidate parameters $\hat{\theta}\seq{l}\interv{\seq{x_1^N}_{\hat{\bsz}=k}^d}$. Then, for each dimension $d$, choose an element $\hat{\theta}_{k,d}$ from $\set{\hat{\theta}\seq{1}\interv{\seq{x_1^N}_{\hat{\bsz}=k}^d},\cdots,\hat{\theta}\seq{L}\interv{\seq{x_1^N}_{\hat{\bsz}=k}^d}}$ by applying $\blueC{\Delta}^1$ to the samples $\seq{x_1^N}_{\hat{\bsz}=k}^d$.
        This step corresponds to the Algorithm \ref{alg:MarginEstimation}.
        \item For each $m\in\set{1,\cdots,M}$, fit the copula form $G_l$ to $\seq{\x_1^N}_{\hat{\bsz}=k}$ by predefined copula estimator, and get the candidate copula parameters $\hat{\alpha}\seq{m}\interv{\seq{\x_1^N}_{\hat{\bsz}=k}}$. Then choose an element $\hat{\alpha}_k$ from $\set{\hat{\alpha}\seq{1}\interv{\seq{\x_1^N}_{\hat{\bsz}=k}},\cdots,\hat{\alpha}\seq{M}\interv{\seq{\x_1^N}_{\hat{\bsz}=k}}}$ by applying $\blueC{\Delta}^2$ to the samples $\seq{\x_1^N}_{\hat{\bsz}=k}$.
        This step corresponds to the Algorithm \ref{alg:CopulaEstimation}.
    \end{enumerate}
    \item Update the parameters $\bsTheta^{(i)}$ by $\set{\hat{\pi}_k, \hat{\alpha}_k, \hat{\bstheta}_k}_{k=1}^K$, where $\hat{\bstheta}_k=\set{\hat{\theta}_{k,d}}_{d=1}^D$.
\end{enumerate}
The above steps are iterated $iterMax$ times until convergence. The \blueCN{whole procedure is summarized} in Algorithm \ref{alg:ICE for CBMMs}.

\IncMargin{1em}
\begin{algorithm}[t]
    \caption{GICE for CBMMs}
    \label{alg:ICE for CBMMs}
    \SetKwInOut{Input}{input}\SetKwInOut{Output}{output}
    \SetKwFunction{Initialization}{Initialization}
    \SetKwFunction{MarginEstimation}{MarginEstimation}
    \SetKwFunction{CopulaEstimation}{CopulaEstimation}
    \Input{$\xunN$, $K$, $iterMax$, \blueC{$T$}, $\H$, $\G$, $\{\hat{\theta}\seq{l}\}_{l=1}^L$, $\set{\hat{\alpha}\seq{m}}_{m=1}^M$, $\blueC{\Delta}^1$, $\blueC{\Delta}^2$}
    \Output{$\bsTheta^{iterMax}$}
    \BlankLine
    $\bsTheta^{(0)}$ = \Initialization{$\xunN$, $K$}\;
    \For{$i\leftarrow1$ \KwTo $iterMax$}{
    simulate \blueC{$\hat{\bsz} = \seq{\hat{\bsz}^1,\cdots,\hat{\bsz}^T}$} according to \eqref{eq:GICE posterior of z} using $\bsTheta^{(i)}$\;
    \For{$k\leftarrow1$ \KwTo $K$}{
        get subgroup sample $\seq{x_1^N}_{\hat{\bsz}=k}$ according to \eqref{eq:GICE subgroup samples}\;
        update $\hat{\pi}_k$ by \eqref{eq:GICE update pi_k}\;
        $\hat{\alpha}_k$ = \MarginEstimation{$\seq{\x_1^N}_{\hat{\bsz}=k}$, $\H$, $\{\hat{\theta}\seq{l}\}_{l=1}^L$, $\blueC{\Delta}^1$}\;
        $\hat{\bstheta}_k$ = \CopulaEstimation{$\seq{\x_1^N}_{\hat{\bsz}=k}$, $\G$, $\set{\hat{\alpha}\seq{m}}_{m=1}^M$, $\blueC{\Delta}^2$}\; 
        }
    $\bsTheta^{(i)}\leftarrow$ $\set{\hat{\pi}_k,\hat{\alpha}_k,\hat{\bstheta}_k}_{k=1}^K$ 
    }
\end{algorithm}
\DecMargin{1em} 

\IncMargin{1em}
\begin{algorithm}[t]
    \caption{Margin Estimation}
    \label{alg:MarginEstimation}
    \SetKwInOut{Input}{input}\SetKwInOut{Output}{output}
    \Input{$\seq{\x_1^N}_{\hat{\bsz}=k}$, $\H$, $\{\hat{\theta}\seq{l}\}_{l=1}^L$, $\blueC{\Delta}^1$}
    \Output{$\hat{\bstheta}_k$}
    \For{$d=1\leftarrow$ \KwTo $D$}{
        $\theta_{tmp}\leftarrow\set{\ }$\;
        \For{$l=1\leftarrow$ \KwTo $L$}{
            $\theta_{tmp} \leftarrow \theta_{tmp} + \hat{\theta}\seq{l}\interv{\seq{x_1^N}_{\hat{\bsz}=k}^d}$
        }
        $\hat{\theta}_{k,d} \leftarrow$ the unique element $\in\theta_{tmp}$ decided by $\blueC{\Delta}^1$
    }
\end{algorithm}
\DecMargin{1em} 

\begin{algorithm}[htbp]
    \caption{Copula Estimation}
    \label{alg:CopulaEstimation}
    \SetKwInOut{Input}{input}\SetKwInOut{Output}{output}
    \Input{$\seq{\x_1^N}_{\hat{\bsz}=k}$, $\G$, $\set{\hat{\alpha}\seq{m}}_{m=1}^M$, $\blueC{\Delta}^2$}
    \Output{$\hat{\alpha}_k$}
    $\alpha_{tmp}\leftarrow\set{\ }$\;
    \For{$\blueC{m}=1\leftarrow$ \KwTo $\blueC{M}$}{
        $\alpha_{tmp} \leftarrow \alpha_{tmp} + \hat{\alpha}\seq{m}\interv{\seq{\x_1^N}_{\hat{\bsz}=k}}$
    }
    $\hat{\alpha}_k\leftarrow$ the unique element $\in\alpha_{tmp}$ decided by $\blueC{\Delta}^2$
\end{algorithm}

\subsection{Computational details}
\label{sec:Computational details}

\subsubsection{Initialization of parameters}
\label{sec:Initialization of parameters}

There are various techniques to initialize the parameters $\bsTheta^{0}$ corresponding to \blueCN{S}tep 1 in \blueCN{A}lgorithm \ref{alg:ICE for CBMMs}. \blueC{C}onsidering methods with hard assignment, one can simulate the hidden labels $\hat{\bsz}$ randomly or through K-Means, 
then run \blueCN{S}teps 4-9 to obtain the $\bsTheta^{0}$. Considering soft assignment, simpler mixtures can be applied, and the learnt parameters of the mixture are taken as $\bsTheta^{0}$. 
For instance, using GMM and converting the learnt means and covariances to marginal and copula parameters. K-Means and GMM initialization are both tested in this work, as demonstrated in \blueC{S}ection \ref{sec:Test on simulated data of Non-Gaussian CBMM}.

\subsubsection{Estimators and decision rules}
\label{sec:Estimators and decision rules}
The ICE framework is general and allows different choices of estimators and decision rules for assumptions 1-2. For marginal distributions we can use the Maximum Likelihood \blueCN{Estimator (MLE)}:
\begin{equation}
    \label{eq:marginal estimator}
    \hat{\theta}\seq{l}=\argmax{\blueC{\theta\seq{l}}}\interv{\sum_{q=1}^Q \Ln \seq{f_{\theta\seq{l}}\seq{y_q}}},
\end{equation}
\blueCN{Alternatively, one could also use the method of moments when the sample size of the clusters is large} \cite{fujimaki2011online}. Similarly for copulas, parameters can be estimated through ML:
\begin{equation}
    \label{eq:copula estimator}
    \hat{\alpha}\seq{m}=\argmax{\blueC{\alpha\seq{m}}}\interv{\sum_{q=1}^Q\Ln\seq{ c_{\alpha\seq{m}}\seq{\hat{\u}_q}}},
\end{equation}
where $\hat{\u}_q = \seq{\hat{u}_q^1,\cdots,\hat{u}_q^D}$ represents the pseudo samples of the copula which are the estimated marginal CDFs of samples $\set{y_q^1,\cdots,y_q^D}$. There are basically two ways to get these pseudo samples: (i) the parametric way, using the estimated parameters of marginal distributions to compute the $\hat{\u}$ \blueC{through their CDF formulas}, and (ii) the non-parametric way, estimating $\hat{\u}$ empirically \blueC{by $\hat{F}(y) = \frac{1}{Q}\sum_{q=1}^{Q}\mathbb{1}_{Y \leq y}$}. \blueC{The ML estimator of copula \eqref{eq:copula estimator} taking $\hat{\u}$ generated parametrically refers to the Inference Function for Margin (IFM) method \cite{joe1996estimation}, while the estimator taking  $\hat{\u}$ generated empirically is called Pseudo Maximum Likelihood (PML) method  \cite{genest1995semiparametric}}\blueCN{.}
Alternatively, copula parameters can also be estimated through the method of moments by means
of the empirical estimation of Kendall’s tau \cite{kendall1938new}. 
In this work, ML is chosen as marginal estimator and PML is chosen as copula estimator for all candidate forms. \blueCN{When} using the method of moments approach, one should pay attention to the chosen order of moments, and the relation between parameters and moments which may not exist in close form for some candidate forms.

The minimization of the Kolmogorov distance is adopted as decision rule for both $\blueC{\Delta}^1$ and $\blueC{\Delta}^2$:
\begin{align}
    \label{eq:D1}
    \blueC{\Delta}^1\seq{y_1^Q} &= \arginf{l\in\set{1,\cdots,L}}\interv{\argsup{y\in\set{y_1^d,\cdots,y_Q^d}}\left| F^d\seq{y} - F^d_{\hat{\theta}\seq{l}}\seq{y}\right|}, \\
    \label{eq:D2}
    \blueC{\Delta}^2\seq{\y_1^Q} &= \arginf{m\in\set{1,\cdots,M}}\interv{\argsup{\y\in\set{\y_1,\cdots,\y_Q}}\left| F\seq{\y} - F_{\hat{\alpha}\seq{m}}\seq{\y}\right|},
\end{align}
where $F^d\seq{y}$ and $F\seq{\y}$ are approximated by empirical CDFs. Other possible decision rules can be ML (PML for $\blueC{\Delta}^2$ \cite{derrode2016unsupervised}), or some particular rules for specific families of candidate forms. For example, $\blueC{\Delta}^1$ can be based on ``skewness'' and ``kurtosis'' if all candidate marginal forms belong to the Pearson family \cite{delignon1997estimation}.

\subsection{Convergence}
\label{sec:Convergence}
It is difficult to study theoretically the convergence of GICE, as the algorithm allows different combinations of estimators and decision rules. Nevertheless, 
it is
easy to trace a certain convergence index at each iteration in order to monitor the performance. For instance, the Kolmogorov distance between empirical CDFs and the learnt CBMM CDFs of the given data:
\begin{equation}
    \label{eq:convergence index}
    d\seq{F,F_{\hat{\bsTheta}}} = \argsup{\x\in\set{\x_1,\cdots,\x_N}}\left| F\seq{\x} - F_{\hat{\bsTheta}}\seq{\x}\right|,
\end{equation}
where $F_{\hat{\bsTheta}}\seq{\x}=\sum_{k=1}^K \hat{\pi}_k F_{\hat{\alpha}_k,\hat{\bstheta}_k}\seq{\x}$.
\blueCN{In practice, one should choose a large enough $iterMax$ to achieve convergence, or stop the iterations when a chosen criterion is met. For example, when no} change of the estimated marginal and copula forms is observed between two iterations, and the difference of the estimated parameters is within some predefined threshold. 

\blueC{Besides the $iterMax$ parameter, the realisation time parameter $T$ also plays an important role on the convergence of GICE by affecting its smoothness. Increasing $T$ increases the size of simulated samples, based on which the distribution forms and parameters are estimated. Thus, choosing a larger $T$ leads to a smoother path of convergence. Further discussion on $T$ can be found with in the synthetic experiments (Section \ref{sec:Test on simulated data of Non-Gaussian CBMM}).}

\section{Experiments and results}
\label{sec:Experiments}

\blueC{W}e consider data dimension $D=2$ and the following candidate forms for all the experiments:
\begin{itemize}
    \item[] $\H = \set{\text{Gamma, Fisk, Gaussian, T, Laplace, Beta, BetaPrime}}$,
    \item[] $\G = \set{\text{Gumble, Gaussian, Clayton, FGM, Arch12, Arch14, Product}}$.
\end{itemize}
Their PDF, CDF, and parameter details are listed in Appendix A.

\subsection{Experiments on synthetic data}
\label{sec:Experimentation on synthetic data}

This series of experiments aims to test the efficiency of the GICE for the identification of CBMMs constructed by different marginal and copula forms\blueC{, with component number $K=2$}. Some computational details that impact the performance of GICE are also illustrated and discussed in this section.

\subsubsection{Test on simulated data of Non-Gaussian CBMM}
\label{sec:Test on simulated data of Non-Gaussian CBMM}
\begin{figure}[tb]
\centering
\begin{subfigure}[t]{0.49\textwidth}
  \centering
  \includegraphics[width=0.95\textwidth]{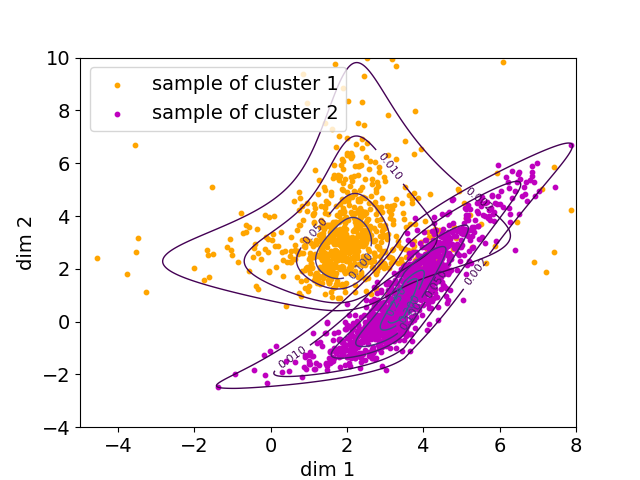}
  \captionsetup{width=.9\linewidth}
  \caption{True cluster densities and labels, CBMM parameters as listed in Table \ref{tab:nonGauss True and estimated CBMM parameters}}
  \label{fig:nonGauss True PDF and labels}
\end{subfigure}%
\begin{subfigure}[t]{0.49\textwidth} 
  \centering
  \includegraphics[width=0.95\textwidth]{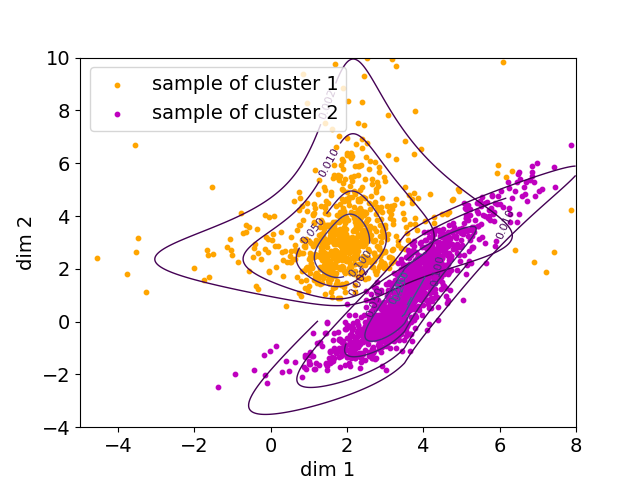}
  \captionsetup{width=.95\linewidth}
  \caption{Estimated cluster densities and labels (GICE \blueC{with realization time $T$}=10, GMM initialization, \blueC{100 max. iterations}).}
  \label{fig:nonGauss Estimated PDF and labels}
\end{subfigure}
\caption{Synthetic experiment \blueCN{(N=2000 samples)} to evaluate the performance of GICE on non-Gaussian CBMM identification.}
\label{fig:nonGauss CBMM PDF and labels}
\end{figure}

This series of tests aims to verify the performance of GICE on non-Gaussian CBMM identification. 2000 samples were simulated from a pre-defined CBMM with non-Gaussian marginal and copulas. The model parameters are listed in the second row of Table \ref{tab:nonGauss True and estimated CBMM parameters}. Figure \ref{fig:nonGauss True PDF and labels} shows the simulated data and the cluster densities. The two clusters 
were designed to partially overlap and imitate a difficult identification situation, since clearly separated clusters are much easier to identify, and even simply using K-Means may well predict the labels.

We tested two different initialization methods: the K-Means and the EM for GMM, and considered different realization times $T=1$ and $T=10$. 
We set a maximum of $100$ iterations for GICE to reach its convergence. The evolution of the Kolmogrorov distance between the learnt model and the empirical data distribution  with iterations is reported in 
Appendix B, Figure \ref{fig:nonGauss gof}. 
The evolution of error ratio between the predicted labels and the true ones with iterations is illustrated in 
Appendix B, Figure \ref{fig:nonGauss error ratio}. 
For these synthetic data generated from the defined CBMM as in Figure \ref{fig:nonGauss True PDF and labels}, we see that EM for GMM provides a better-fit initialization than K-Means, by comparing the 
Kolmogorov distance and error ratio of the two GICE settings ``$T=10$, init: K-Means'' and ``$T=10$, init: GMM'' \greenC{at the start of the algorithm}
in Figure \ref{fig:nonGauss CBMM convergence}. However, after 30 iterations they converged to the same level and give very similar estimated CBMMs in the end, see the estimated marginal and copula distributions in 
the third and fourth rows of Table \ref{tab:nonGauss True and estimated CBMM parameters}. 

\begin{table}[b]
\caption{True and estimated CBMM parameters with different realization times and initializations. \blueC{$T$: realization time, init: initialization method}.}
\label{tab:nonGauss True and estimated CBMM parameters}
\centering
    \resizebox{1\columnwidth}{!}{
        \begin{tabular}{c||c|c|c|c||c|c|c|c|}
            \cline{2-9}
             & $\pi_{k=1}$ & $\theta_{k=1,d=1}$ & $\theta_{k=1,d=2}$ & $\alpha_1$ & $\pi_{k=2}$ & $\theta_{k=2,d=1}$ & $\theta_{k=2,d=2}$ & $\alpha_2$ \\
            \hline
            True $\bsTheta$ & 0.4 & \makecell{T \\(2, 2, 0.7)}  & \makecell{Fisk \\(4, 0, 3)} &  \makecell{FGM \\(1)}& 0.6 & \makecell{Laplace \\(3.5, 0.8)} & \makecell{ Gamma\\(10, -4, 0.5)} & \makecell{Arch14 \\(3)} \\
            \hline
            \makecell{GICE $T$=10 \\ init: K-Means} & 0.38 & \makecell{T \\(1.74, 1.95, 0.64)}  & \makecell{Fisk \\(3.78, 0.29, 2.77)} & \makecell{FGM \\(0.89)} & 0.62 & \makecell{Laplace \\(3.51, 0.79)}  & \makecell{T \\(27.0, 1.00, 1.49)} & \makecell{Arch14 \\(2.95)} \\
            \hline
            \makecell{GICE $T$=10 \\ init: GMM} & 0.38 & \makecell{T \\(1.76, 1.95, 0.65)}  & \makecell{Fisk \\(3.80, 0.28, 2.78)} & \makecell{FGM \\(0.92)} & 0.62 & \makecell{Laplace \\(3.51, 0.79)}  & \makecell{T \\(24.6, 1.00, 1.49)} & \makecell{Arch14 \\(2.95)} \\
            \hline    
            \makecell{GICE $T$=1 \\ init: GMM} & 0.38 & \makecell{T \\(1.73, 1.94, 0.63)}  & \makecell{Fisk \\(3.84, 0.22, 2.84)} & \makecell{FGM \\(0.94)} & 0.62 & \makecell{Laplace \\(3.51, 0.79)}  & \makecell{T \\(37.4, 1.02, 1.51)} & \makecell{Arch14 \\(2.91)} \\
            \hline 
        \end{tabular}
    }
\end{table}

GICE is a stochastic method, thus the curves of the convergence indexes are not smoothly 
decreasing.
The change of the selected distribution forms between iterations may lead to small burrs in these curves. In spite of the stochastic nature of GICE, the realization time $T$ also impacts the smoothness of the convergence. The curves obtained from the settings ``$T=10$, init: GMM'' are smoother than the ones obtained from ``$T=1$, init: GMM'' in both Figure \ref{fig:nonGauss gof} and Figure \ref{fig:nonGauss error ratio}. The estimated parameters with only one realization are not too far from the result of ten realizations as reported in the last row of Table \ref{tab:nonGauss True and estimated CBMM parameters}. 
Although increasing $T$ reduces the oscillation around the target parameters at convergence, it also means to increase the computational time of GICE. In this experiment, GICE took 207 s when $T=1$, and 1548 s when $T=10$, running on a 2.7MHz CPU. In practice, $T$ should be chosen such that it balances the expected convergence smoothness and time cost.

Figure \ref{fig:nonGauss Estimated PDF and labels} shows the identified cluster densities and predicted labels through GICE with $T=10$ realizations and GMM initialization. GICE attempts to find the best-fit CBMM for the data, but it does not ensure to always find the true marginal and copula forms. Many factors have impact on its final form decision. For example, the estimation condition, including the 
number of available samples and the difficulty to separate the clusters. This situation is not something we can improve. But in practice, by carefully examining the GICE convergence, we can try different candidate forms, initialization methods, estimators, and run GICE with different random seeds, to avoid being trapped in local extrema, and increase the chance of finding the best-fit model. 

\begin{figure}[tb]
\centering
\begin{subfigure}[t]{0.49\textwidth}
  \centering
  \includegraphics[width=0.95\textwidth]{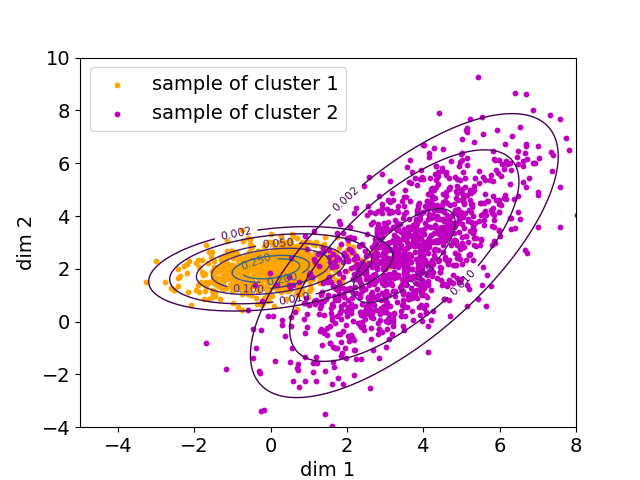}
  \captionsetup{width=.8\linewidth}
  \caption{True cluster densities and labels, GMM parameters as listed in Table \ref{tab:Gauss True and estimated CBMM parameters}}
  \label{fig:Gauss True PDF and labels}
\end{subfigure}%
\begin{subfigure}[t]{0.49\textwidth}
  \centering
  \includegraphics[width=0.95\textwidth]{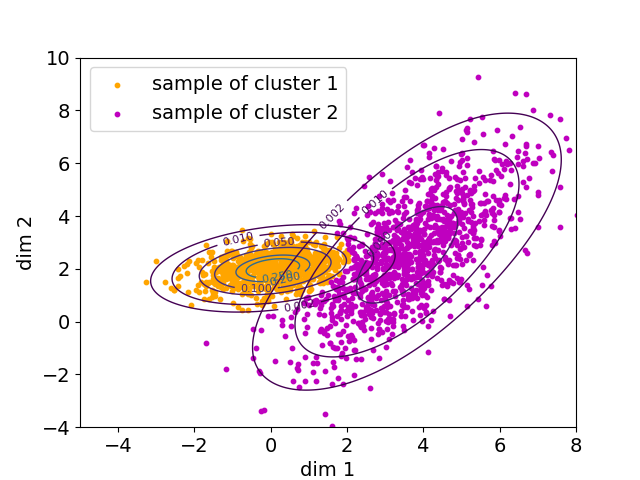}
  \captionsetup{width=.95\linewidth}
  \caption{Estimated cluster densities and labels (GICE \blueC{with realization time $T$}=10, K-Means initialization, \blueC{100 max. iterations}).}
  \label{fig:Gauss Estimated PDF and labels}
\end{subfigure}
\caption{Synthetic experiment \blueCN{(N=2000 samples)} to evaluate the performance of GICE on GMM identification.}
\label{fig:Gauss CBMM PDF and labels}
\end{figure}

\begin{table}[b]
\caption{True Gaussian CBMM (GMM) and estimated parameters with different initializations. \blueC{$T$: realization time, init: initialization method}.}
\label{tab:Gauss True and estimated CBMM parameters}
\centering
    \resizebox{1\columnwidth}{!}{
        \begin{tabular}{c||c|c|c|c||c|c|c|c|}
            \cline{2-9}
             & $\pi_{k=1}$ & $\theta_{k=1,d=1}$ & $\theta_{k=1,d=2}$ & $\alpha_1$ & $\pi_{k=2}$ & $\theta_{k=2,d=1}$ & $\theta_{k=2,d=2}$ & $\alpha_2$ \\
            \hline
            True $\bsTheta$ & 0.4 & \makecell{Gaussian \\(0, 1)}  & \makecell{Gaussian \\(2, 0.5)} & \makecell{Gaussian \\ 0.3} & 0.6 & \makecell{Gaussian \\(3.5, 1.5)} & \makecell{Gaussian \\(2.5, 2)} & \makecell{Gaussian \\0.7} \\
            \hline
            \makecell{GICE $T$=10 \\ init: \blueC{K-Means}} & 0.42 & \makecell{T \\(195, 0.07, 1.01)}  & \makecell{Gaussian \\(2.02, 0.52)} & \makecell{Gaussian \\(0.33)} & 0.58 & \makecell{T \\(28.2, 3.60, 1.42)}  & \makecell{T \\(39.8, 2.57, 1.92)} & \makecell{Gaussian \\(0.69)} \\
            \hline
            \makecell{GICE $T$=10 \\ init: GMM} & 0.42 & \makecell{T \\(29.6, 0.06, 0.98)}  & \makecell{Gaussian \\(2.02, 0.51)} & \makecell{Gaussian \\(0.27)} & 0.58 & \makecell{T \\(28.2, 3.59, 1.42)}  & \makecell{T \\(39.0, 2.57, 1.92)} & \makecell{Gaussian \\(0.69)} \\
            \hline    
        \end{tabular}
    }
\end{table}

\subsubsection{Test on simulated data of GMM}
This series of tests aims to check if GICE also works
for the identification of GMM, which can be considered a special case of CBMM.

2000 samples were simulated from a pre-defined two-component CBMM with Gaussian \blueCN{marginals} and copulas (thus, it is actually a GMM), see Figure \ref{fig:Gauss True PDF and labels} for the simulated data and the true cluster densities. The parameters are reported in the first row of Table \ref{tab:Gauss True and estimated CBMM parameters}. 
$T=10$ was set as realization time for GICE, and similar to the previous series, we 
tested two initialization methods: K-Means and EM for GMM. 
Initializing by EM for GMM means to start with the optimized parameters, and GICE does not drive the estimated model far away from the already optimized one, as we observe stable horizontal lines for both the evolution of the Kolmogorov distance and the error ratio in Figure \ref{fig:GMM convergence}. Meanwhile, GICE converges to the same level with K-Means initialization, and the identified model is quite close to the one initialized by EM for GMM, see the \blueC{second} and \blueC{third} rows in Table \ref{tab:Gauss True and estimated CBMM parameters}. It is reasonable that 
the T distribution is often selected by GICE instead of the Gaussian (the ground truth),
because Gaussian and T distributions with an infinite amount of degrees of freedom
are identical, and the estimated parameters of T distributions are all with relatively large degrees of freedom.
Comparing the two sub-figures in Figure \ref{fig:Gauss CBMM PDF and labels}, differences are hardly noticeable visually between the identified cluster density through GICE and the ground truth.

\subsection{\blueC{Experiments on real data: MNIST database}}
\label{subsec:MNIST}

\blueC{This experiment expands the previous evaluation of GICE to 
a real dataset of more than two clusters ($K=10$).
We tested our method on the entire MNIST database \cite{Lecun:1998} \greenC{(N=70000 samples)}, which consists of $28 \times 28$ pixels grayscale images of handwritten digits from $0$ to $9$, therefore consisting of 10 classes.}

To minimize the challenges related to analyzing such high-dimensional data, \blueC{the Uniform Manifold Approximation and Projection (UMAP) algorithm \cite{mcinnes2018umap} was used as a preprocessing step to project data into a 2D space before applying the clustering methods. UMAP can preserve both the local and global structure of data, and improves the performance of downstream clustering methods on the MNIST database, as already illustrated in several published studies. For example, \cite{allaoui2020considerably} reports the improvements of three UMAP preprocessed clustering algorithms K-Means, HDBSCAN \cite{campello2013density} and GMM. \cite{mcconville2021n2d} layers up UMAP, auto-encoder and GMM into a method called ``N2D'', and achieves a clustering accuracy of 0.979 on MNIST.} 
UMAP is sensitive to its parameter K-Nearest Neighbors (KNN), which controls the balance of local versus global structure in the data. A large KNN risks to merge the real clusters together, while a small one may lead to many subgroups of samples in the projected space. 
\blueC{In this experiment, KNN$=30$ was chosen as suggested by the documentation of UMAP to avoid producing fine grained cluster structure that may be more a result of noise patterns.}

\begin{figure}[t]
\centering
\begin{subfigure}[t]{0.5\textwidth}
  \centering
  \includegraphics[width=0.815\textwidth]{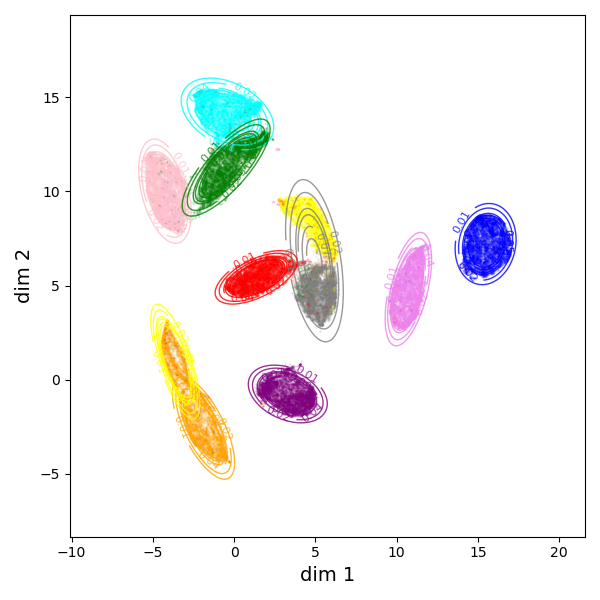}
  \captionsetup{width=.9\linewidth}
  \caption{\blueC{Cluster densities found by GMM-EM, accuracy=0.825, Kolmogorov distance=0.029.}}
  \label{fig:MNIST density GMM}
\end{subfigure}%
\begin{subfigure}[t]{0.5\textwidth}
  \centering
  \includegraphics[width=0.95\textwidth]{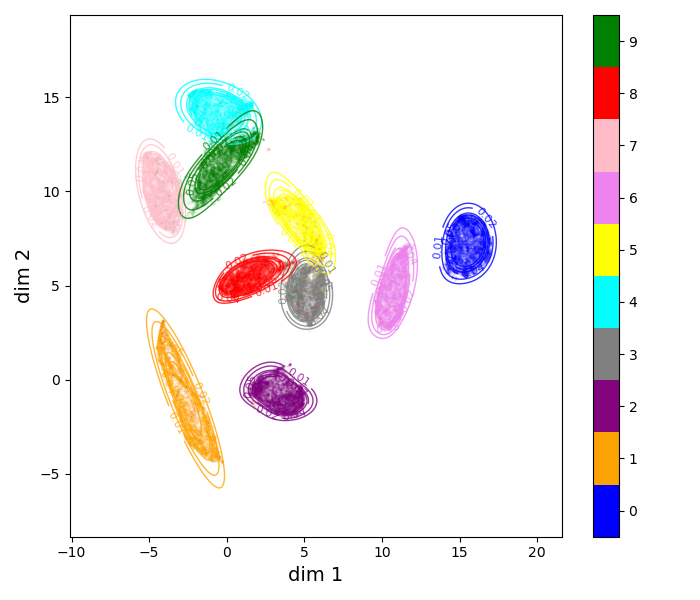}
  \captionsetup{width=.95\linewidth}
  \caption{\blueC{Cluster densities found by CBMM-GICE, accuracy=0.967, Kolmogorov distance=0.011.}}
  \label{fig:MNIST density CBMM}
\end{subfigure}
\caption{\blueC{Experiment on the MNIST dataset (\greenC{N=70000 samples, }10 clusters, 2D projection obtained by UMAP with KNN=30), to evaluate the performance of GICE on more than two clusters. Each point corresponds to the image of a digit, colored by its corresponding ground truth label.}}
\label{fig:MNIST density estimation}
\end{figure}

\begin{table}[b]
    \centering
    \resizebox{0.7\columnwidth}{!}{
    \begin{tabular}{c||c|c|c||c|c|c|}
        \cline{2-7}
        & \multicolumn{3}{c||}{\blueC{Accuracy}} & \multicolumn{3}{|c|}{\blueC{Kolmogorov distance}} \\
        \cline{2-7}
        & \blueC{average} & \blueC{min} & \blueC{max} & \blueC{average} & \blueC{min} & \blueC{max} \\
        \hline
        \blueC{GMM-EN} & \blueC{0.824} & \blueC{0.690} & \blueC{0.956} & \blueC{0.025} & \blueC{0.013} & \blueC{0.055}\\
        \hline
        \blueC{CBMM-GICE} & \blueC{0.848} & \blueC{0.716} & \blueC{0.967} & \blueC{0.021} & \blueC{0.011} & \blueC{0.029}\\
        \hline
    \end{tabular}}
    \caption{\blueC{Performance of GMM-EM and CBMM-GICE on the MNIST dataset based on 20 repeated experimentation (100 max iterations for GMM and GICE, GICE realization time $T$=10, initialization: GMM).}}
    \label{tab:Performance of GMM-EM and CBMM-GICE on MNIST dataset}
\end{table}

\blueC{Table \ref{tab:Performance of GMM-EM and CBMM-GICE on MNIST dataset} reports the clustering results of 20 repeated experiments on the projected (2D) MNIST data through CBMM-GICE and GMM-EM. For easy comparison to the existing results provided in other studies, we report the accuracy as measure of performance instead of the error ratio we used in previous sections. 
We observe that both algorithms manage to identify the relevant clusters, whereas non-Gaussian distributions are slightly more desirable with better goodness of fit. The GMM-EM performance is quite close to the one (0.825) reported in \cite{mcconville2021n2d}. 
Figure \ref{fig:MNIST density estimation} shows the \blueC{2D} latent space estimated by UMAP, corresponding to the best result of GICE-CBMM in the table. Each point corresponds to the image of a digit, colored by its corresponding ground truth label. Most of the clusters are already separated by UMAP, their tight distribution offers a good condition for GMM estimation. Dealing with the overlapping clusters is more challenging, and in this experiment, GMM mistook the two clusters in the center as one entire Gaussian, and force-created two subgroups in the cluster at the lower-left corner, as illustrated in Figure \ref{fig:MNIST density GMM}.}


\subsection{Experiments on real data: subgroup clustering for the characterization of myocardial infarct patterns}
\label{subsec:Application to subgroup clustering for myocardial infarct}

\blueCN{In the previous sections, t}he increased flexibility of CBMM and the identification ability of GICE have been tested on simulated data \blueCN{and on a well-known large real dataset}. We now illustrate their relevance on a real data application in cardiac imaging, namely clustering subgroups of patients based on their infarct patterns, previously extracted from medical images.

Myocardial Infarction (MI) is a severe coronary artery disease. It is caused by a shortage of blood flow to a portion of the heart, leading to necrosis of the heart muscle \blueC{\cite{Ibanez:EHJ:2018}}. Imaging techniques play a crucial role for MI diagnosis and the evaluation of infarct evolution after reperfusion.
However, infarct assessment as performed in clinical routine relies on oversimplified scalar measurements of the lesions, mostly infarct extent and transmurality (the amount of propagation across the myocardial wall, from endocardium to epicardium). Infarct patterns are much more complex, and require more advanced analysis tools to characterize their \blueC{severity} across a population and potential evolution \cite{Duchateau:FrontCV:20233}. Here, we aim at demonstrating the relevance of the CBMM identified by GICE, simply called CBMM-GICE later, for two clustering problems: (i) separating infarcts from different territories, (ii) identifying subgroups of infarct patterns \blueC{for a given territory and examine their severity}. \blueCN{Given the complexity of infarct shapes, subgroups of patients are very likely to follow a non-Gaussian distribution, which can be critical for any subsequent statistical analysis that compares subgroups of patients or an individual to a given subgroup.}

We focus on the imaging data from the HIBISCUS-STEMI cohort \blueC{(ClinicalTrials ID: NCT03070496)}, 
which is a prospective cohort of patients with ST Elevation Myocardial Infarction (STEMI). The infarct patterns were manually segmented offline on late Gadolinium enhancement \blueC{magnetic resonance} images using commercial software (CVI42 v.5.1.0 Circle Cardiovascular Imaging, Calgary, Canada). Then, they were resampled to a reference anatomy using a parameterization of the myocardium in each image slice, as done in \cite{Duchateau:FrontCV:20233}. This allows comparing the data from different patients and acquisitions at each voxel location of a common template. After alignment, each subject presents 21 slices ranging from the basal (mitral valve) to apical levels of the left ventricle, \blueC{totalling a number of 8110 voxels within the myocardium (the zone across which we analyze infarct patterns).}

\blueC{In the following experiments, we both examined the distribution of samples in the UMAP 2D latent space, and high-dimensional reconstructions of representative samples. As visualization of infarct patterns across the 21 slices is difficult, we used a polar representation (often called Bull's eye plot) which provides a global view of a given infarct pattern at once. It consists of a flattened version of the left ventricle, made of concentric circles each corresponding to a given slice. The apex and basal levels are at the center and periphery of the Bull's eye. Each circle averages the infarct information across the radial direction of the myocardium (namely, transmurally). This representation is commonly used by clinicians (simplified into 17 segments which can be easily related to their associated coronary territories \cite{CERQUEIRA2002463}, 
see 
Appendix C, Figure \ref{fig:territory}). A summary of these processing and visualization steps is given in 
Appendix C, Figure \ref{fig:infarctDisplay}.}

\subsubsection{Infarct territory clustering}
\label{subsubsec:Infarct territory clustering}

\begin{figure}[tb]
\centering
\begin{subfigure}[t]{0.32\textwidth}
  \centering
  \includegraphics[width=0.95\textwidth]{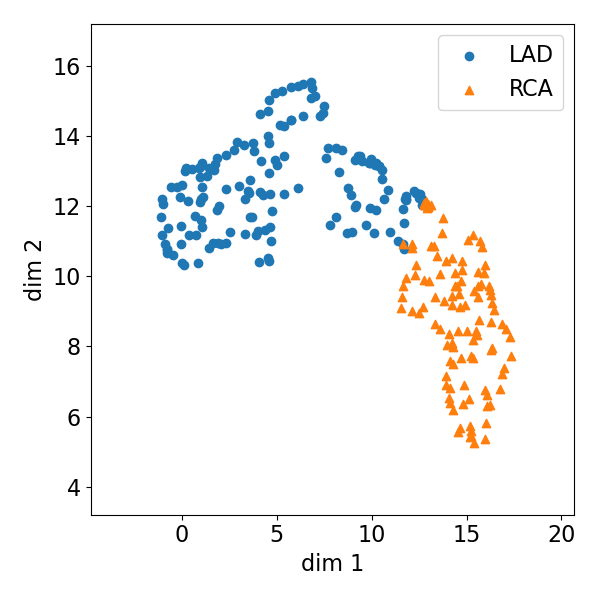}
  \captionsetup{width=.9\linewidth}
  \caption{2D representation of the studied infarct patterns, colored by their coronary territory (LAD or RCA).}
  \label{fig:terryClust_Projected infarct segmentations}
\end{subfigure}%
\begin{subfigure}[t]{0.32\textwidth}
  \centering
  \includegraphics[width=0.95\textwidth]{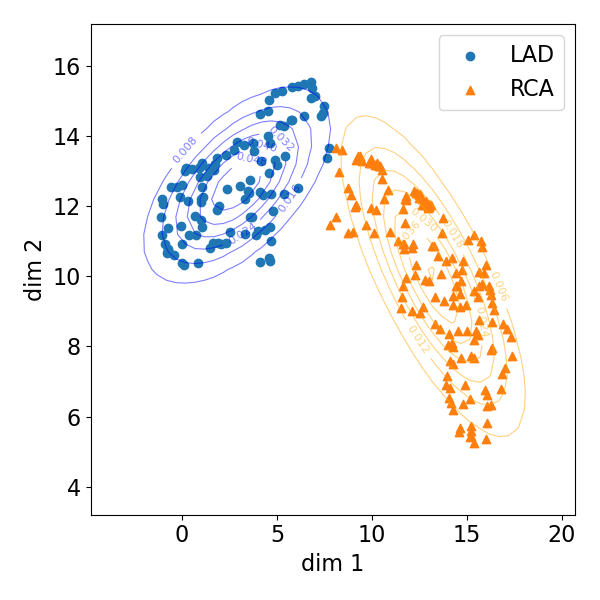}
  \captionsetup{width=.95\linewidth}
  \caption{Clusters and densities found through GMM-EM. \blueC{Both LAD and RCA clusters are Gaussian distributed.}}
  \label{fig:terryClust_result GMM-EM}
\end{subfigure}
\begin{subfigure}[t]{0.32\textwidth}
  \centering
  \includegraphics[width=0.95\textwidth]{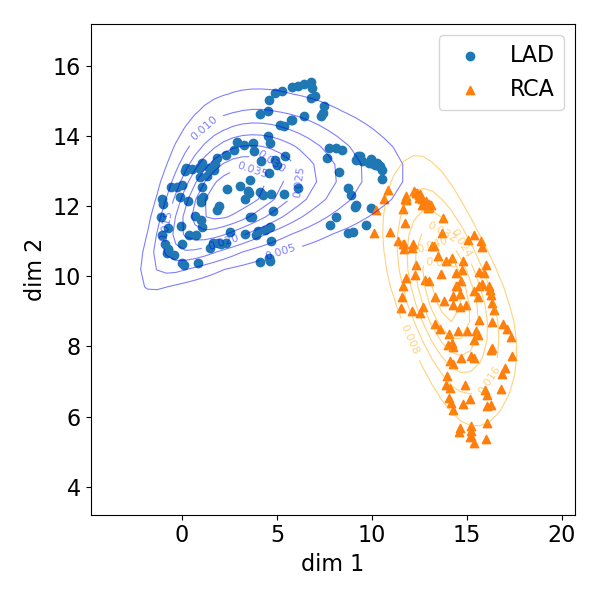}
  \captionsetup{width=.95\linewidth}
  \caption{Clusters and densities found by CBMM-GICE. Cluster LAD: \blueC{Beta prime} and Gamma marginals, Clayton copula. Cluster RCA: Fisk marginals, Gaussian copula.}
  \label{fig:terryClust_result CBMM-GICE}
\end{subfigure} 
\caption{Clustering of the projected infarct segments in LAD and RCA territories \blueCN{(N=276 samples)}. 
The illustrated projection was selected \blueC{using the trustworthiness score \cite{venna2005local} (the highest values corresponding to projections that best preserve local neighborhoods)} among 50 UMAP projections with KNN = 6.
}
\label{fig:terryClust_figures}
\end{figure}

This section deals with the territory clustering task, namely separating patients with infarcts in different territories in the studied dataset through CBMM-GICE. Although this problem could be handled with supervised methods, and is by itself a rather feasible task, we prefer to start by evaluating our methods on this controlled environment before moving to a more challenging problem on the same medical imaging application. \blueCN{Besides, the subgroups of patients are strongly non-Gaussian in this population, as visible in Figure \ref{fig:terryClust_figures}a.}

MI can happen in three territories covering the left ventricle, \blueC{as illustrated in 
Appendix C, Figure \ref{fig:territory}, }corresponding to three primary coronary arteries: the Left Anterior Descending (LAD), the Right Coronary Artery (RCA), and the Left CircumfleX (LCX). In general, the infarcted territories are rather easy to distinguish across patients, as visible in the two clear examples from 
Appendix C, Figure \ref{fig:LAD example} and \ref{fig:RCA example}. Nevertheless, more challenging cases may also be encountered, in particular when the infarct is mostly localized near the apex.

\blueC{In this experiment, w}e used the data from 276 subjects from the HIBISCUS-STEMI cohort, 156 with LAD infarct and 120 with RCA infarct. We did not include patients with LCX infarcts due to the small size of this subgroup.

\blueC{As we didn't find any KNN value for UMAP in related cardiac MRI image studies, we considered various KNN values here, and used the trustworthiness score \cite{venna2005local} to select the best projection for each KNN. This score takes values between 0 and 1, and is a measure of how much local neighborhoods are preserved, which penalizes unexpected nearest neighbors in the output space according to their rank in the input space.} We performed 50 projections and the one with the best trustworthiness score \blueC{was selected} for the downstream clustering. CBMM-GICE was then tested on these selected projections corresponding to different KNN values.

\blueC{Figure \ref{fig:terryClust_figures} shows the 2D latent space estimated by UMAP (each point representing the infarct pattern of a given subject, colored by its corresponding territory label), and the clusters found be GMM-EM and CBMM-GICE.} Comparing the cluster densities obtained by CBMM-GICE in Figure \ref{fig:terryClust_result CBMM-GICE} to the ones obtained by GMM-EM in Figure \ref{fig:terryClust_result GMM-EM}, we observe that without the strict constraint of elliptical form, GICE adapts better to the distribution of projected subjects.

\begin{figure}
    \centering
    \includegraphics[width=.9\textwidth]{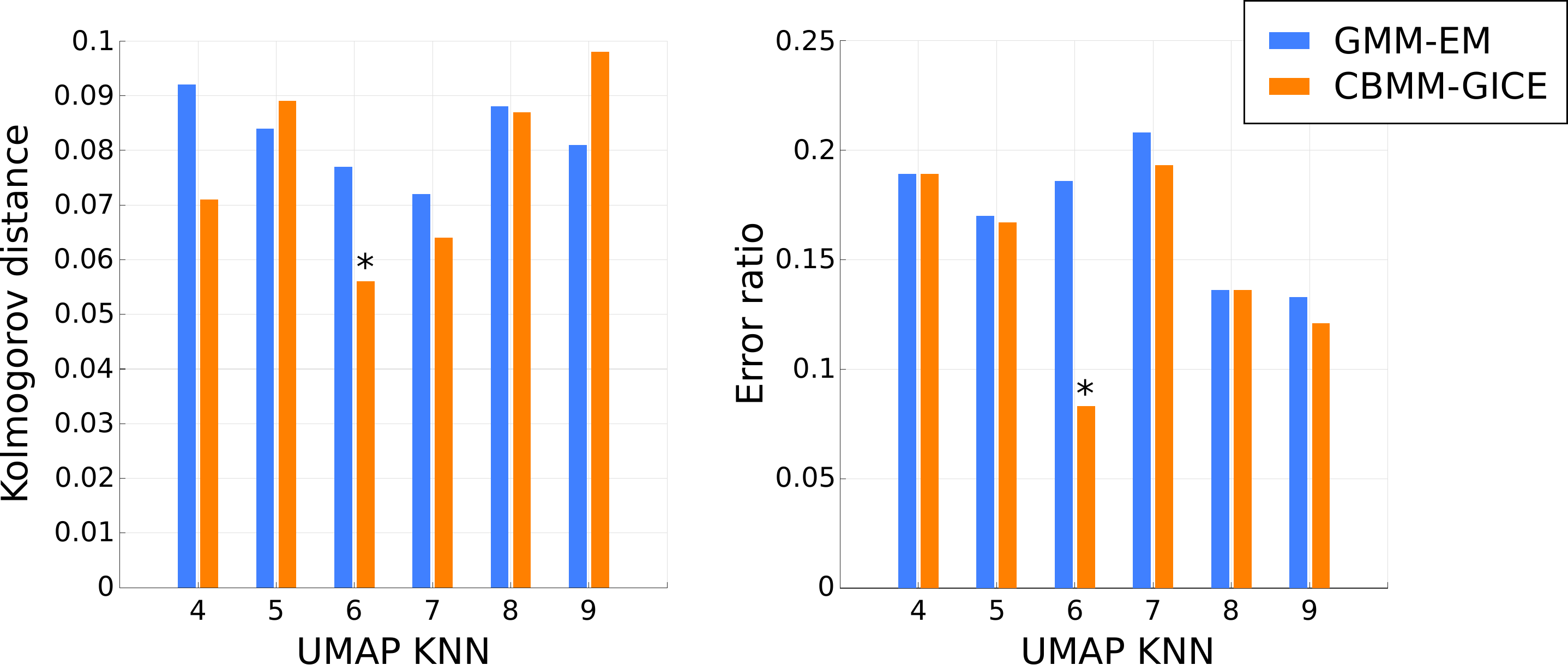}
    \caption{Infarct territory clustering result of GMM-EM and CBMM-GICE (\blueC{realization time} $T$=50, \blueC{GMM initialization}, 100 \blueC{max.} iterations, \blueCN{N=276 samples}). \blueC{The star points out the best result for each metric.}}
    \label{fig:terryClust_table}
\end{figure}

\blueC{Figure \ref{fig:terryClust_table} complements these observations by quantifying} the fitness of estimated CBMM to the distribution of projected subjects, and the error ratio of predicted territory labels, compared to the estimated GMM learnt through EM (GMM-EM). We can conclude that CBMM-GICE performs better than GMM-EM on this territory clustering task. The best clustering result \blueC{was} obtained for KNN=6. 

\subsubsection{Infarct pattern clustering for the LAD territory}
\label{subsubsec:Infarct pattern clustering in LAD territory}

In this section, we test our methods on a more challenging task of clinical relevance, namely identifying subgroups of patients based on their infarct patterns. This is of high clinical value to go beyond the scalar measurements used in clinical practice, and better understand subtle changes in infarct patterns following reperfusion.

We specifically focus on patients whose infarct correspond to the most populated coronary territory in the HIBISCUS-STEMI cohort, namely LAD infarcts. In this population,
54 patients have two visits realized at 1 month and 12 months after their heart has been reperfused. Our objectives are two-fold: finding subgroups of infarct patterns in this specific set of patients, \blueC{potentially of different severity}, and evaluate the migration of patients across clusters between the two visits.

As in the previous Section \ref{subsubsec:Infarct territory clustering}, UMAP was first applied to represent the cohort of LAD infarct segments in to a \blueC{2D} space to ease the mixture model learning and visualization. \blueC{We set KNN=6 as in the previous section.}
Then, the BIC criterion was applied to decide the number of clusters. More precisely, we ran a dozen times the GMM-EM considering different component numbers $K\in\set{1,\cdots,10}$ \blueC{(namely, running the EM algorithm with different GMM parameter initialization)}, and picked the most frequently chosen number according to the BIC criterion. In this way, $K=3$ was selected and considered a suitable cluster number for the LAD infarct segments.
Component number optimization is actually a challenging task in clustering problems but out of the central scope of this study, the alternative methods of BIC can be found in \cite{hancer2017comprehensive}.

\begin{figure}[tb]
\centering
\begin{subfigure}[t]{0.48\textwidth}
  \centering
  \includegraphics[width=\textwidth]{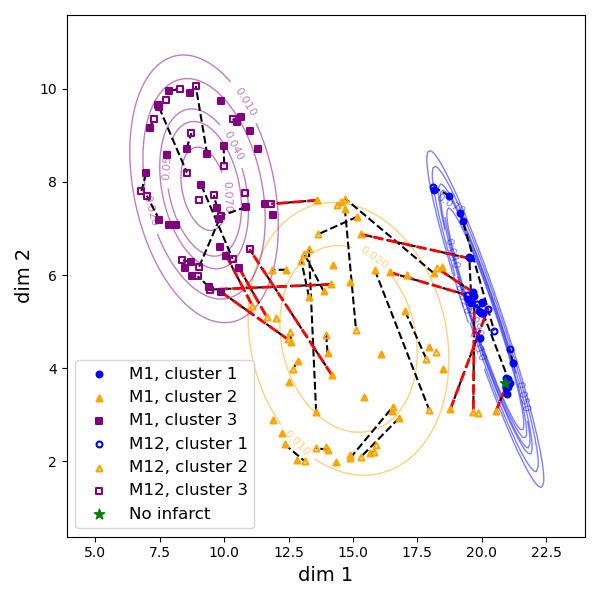}
  \captionsetup{width=.95\linewidth}
  \caption{Estimated cluster densities and labels through by GMM-EM. \blueC{All clusters are Gaussian distributed.}}
  \label{fig:LADclusterGMM}
\end{subfigure} \hfill
\begin{subfigure}[t]{0.48\textwidth}
  \centering
  \includegraphics[width=\textwidth]{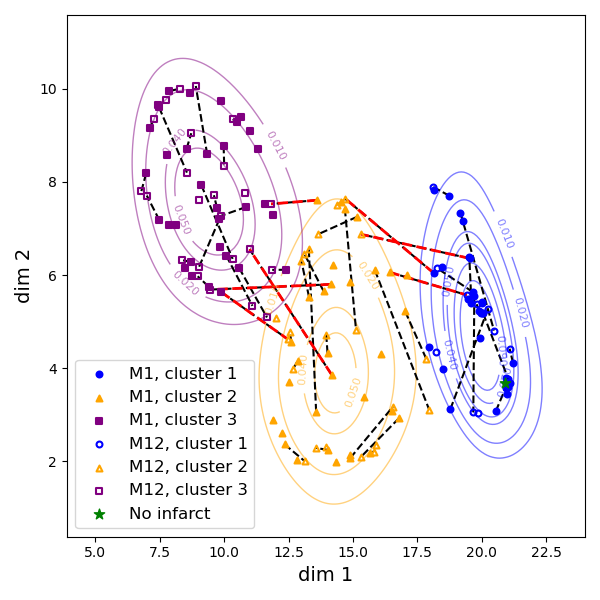}
  \captionsetup{width=.95\linewidth}
  \caption{Estimated cluster densities and labels through CBMM-GICE. \blueC{Cluster 1: Fisk marginals, FGM copula. Cluster 2: Fisk and Beta prime marginals, Gaussian copula. Cluster 3: Gamma and Beta prime marginals, Gaussian copula.}}
  \label{fig:LADclusterCBMM}
\end{subfigure}
\caption{Clustering of LAD infarct patterns (\blueCN{N=54 patients with two visits}, 2D visualization obtained from UMAP with KNN=6). \blueC{Each} dashed line links the two visits of the same patient \blueC{(1 month (M1) and 12 months (M12) after the heart has been reperfused)}.  
In each subplot, the red dots point out cases that migrate from one cluster to another.}
\label{fig:LADClust_figures}
\end{figure}

The CBMM-GICE was tested against the GMM-EM and the MMST-EM (Mixture of Multiple Scaled Student's T identified by EM algorithm \cite{forbes2014new, zheng2020characterization}). The clusters resulting from GMM-EM and CBMM-GICE are illustrated in Figures \ref{fig:LADclusterGMM} and \ref{fig:LADclusterCBMM}, respectively. 
When heavy tail is not present in the clusters, MMST-EM and GMM-EM have very similar performance. 
The clusters obtained through MMST-EM are quite close to the GMM-EM ones, therefore not repeatedly depicted here. 
As no ground truth was available, we computed the mean silhouette scores \cite{rousseeuw1987silhouettes} over all subjects to measure the consistency within the estimated clusters. A high mean silhouette score indicates the subjects well matching 
to their own clusters and poorly matching to the other clusters. 
The clustering performance of all three methods are reported in Table \ref{tab:Clustering result LAD}. Thanks to the flexibility of CBMM, CBMM-GICE not only \blueCN{better fits} to the distribution of subjects as indicated by the Kolmogorov distance, but also results in a highest mean silhouette score of clustering. The resulting clusters are not constrained to be symmetric, they actually follow very different distributions from each other. 

\begin{table}[b]
    \centering
    \resizebox{0.7\columnwidth}{!}{
    \begin{tabular}{c||c|c|c}
        \hline
        Method & GMM-EM & MMST-EM & CBMM-GICE \\
        \hline
        Kolmogorov distance & 0.066 & 0.067 & 0.059 \\
        \hline
        Mean silhouette score & 0.452 & 0.452 & 0.513 \\
        \hline
    \end{tabular}}
    \caption{Evaluation of the clustering quality on patients with LAD infarcts, for GMM-EM, MMST-EM and CBMM-GICE (\blueC{realization time} $T$=50, \blueC{GMM initialization}, 100 \blueC{max.} iterations, \blueCN{N=54 samples}).}
    \label{tab:Clustering result LAD}
\end{table}

To better interpret the \blueC{clinical meaning of the clusters}, we visualized the typical pattern associated with each cluster by reconstructing the high-dimensional infarct pattern corresponding to the cluster ``center''\blueC{, namely }the point with highest probability density. Reconstruction was carried out by multiscale kernel regression \cite{duchateau2013adaptation}, since UMAP is not equipped with an explicit mapping from the low-dimensional representation to the high-dimensional space. 
Appendix C, \blueC{Figure \ref{fig:typical patterns} displays these reconstructed patterns for the CBMM clusters. Comparable results were found for the GMM clusters, given that the cluster centers are rather close between the two methods.}
We observe that clusters not only correspond to infarcts of different extent, but also to complex spreads around the anteroseptal region at the mid-cavity level, and the whole myocardium near the apex (center of the bull's eye).

Differences in the resulting GMM and CBMM clusters are more clearly reflected by examining the evolution of patients between the two visits, depicted with the dash lines in Figure \ref{fig:LADClust_figures}. 
The detected migration of a few patients from one cluster to another are highlighted in red. In spite of the migration commonly detected by GMM-EM and CBMM-GICE, constrained by the symmetric Gaussian form, GMM-EM marks more migrations near the borders of clusters, \blueC{most} of which \blueC{being} being actually with relative short evolution paths.

\section{Conclusion and perspectives}
\label{sec:conclusion}

We adapted the ``Generalized Iterative Conditional Estimation'' (GICE), originally developed for the identification of \blueC{s}witching \blueC{h}idden Markov \blueC{m}odels to the estimation of Copula-Based Mixture Models (CBMM). GICE iteratively optimizes the forms and parameters of the marginals and copulas of the CBMM component distributions driven by data. The resulting CBMM-GICE method allows us to approach any cohort of samples by an appropriate CBMM, and find the subgroups at the same time. Thanks to the flexibility of CBMM, CBMM-GICE generally outperforms the classic mixture models identified through the EM algorithm e.g. GMM-EM, tested both on synthetic and real medical \blueC{imaging} data, where more flexibility in the distribution fit and cluster estimation is often crucial. Aside from this main contribution, this study also tackled the following tasks that have not been addressed in detail before:
\begin{enumerate}
    \item The impact of the realization time $T$ to the GICE algorithm is neglected and always set to 1 in the pioneer works \cite{zheng2020semi,derrode2016unsupervised}. The experiments on synthetic data show that a larger $T$ value can reduce the oscillation around the target CBMM at convergence, especially when facing small sample numbers. However, the choice of $T$ should balance the performance and the computational burden, since increasing $T$ means also increasing the computational time. $T$ around 10 is suggested for a cohort with clusters of around 1000 samples.
    \item The convergence of GICE is hard to prove theoretically. It has only been demonstrated indirectly via the result error ratio in \cite{zheng2020semi}. Nevertheless, we can monitor the convergence through the fitness score of the statistical distributions, such as the Kolmogorov distance applied in the experiments. \blueCN{This manual monitoring may be improved by adding a convergence control module in the algorithm.}
    \item The GMMs are not always a good choice for fitting medical data. 
    In our study, the results of the experiments on cardiac images clustering corroborate that the CBMMs could be the most suitable model, when data subgroups are not elliptically distributed or do not follow the same distribution form (often-met situation). Indeed, GMMs are specific cases of the general CBMMs family when marginals and copulas are all chosen to be Gaussian. 
\end{enumerate}

The \blueC{CBMM-GICE} studied in this article \blueC{has potential}, but still has limits waiting to be improved in the future work. 
As a primary study, the current GICE is developed for \blueC{2D} CBMM only. \blueC{Although the variability of infarct patterns may be contained in slightly more dimensions, 
the present work used 2D visualization to easily focus on the clustering output. Latent spaces of more dimensions will be considered in future work focusing on the clinical application.} The challenge of extending the identification ability of GICE to higher dimension lies in the choice and the estimation of multivariate copulas, since much less multivariate copulas with analytical form are available than bivariate ones. 
\greenC{The vine-copulas or neural-copulas are promising to overcome this limit\cite{sahin2022vine, FAN2024105263}.} More flexibility can be still introduced into the current CBMM. For instance by incorporating the parametric rotation of copulas for component distributions \cite{kosmidis2016model}. Also, 
by jointly estimating a simplified representation of the data and performing clustering, as in deep clustering approaches \cite{Ren_Arxiv:2022}, some of these models reach much better performance than our method on the MNIST dataset\footnote{\url{https://paperswithcode.com/sota/image-clustering-on-mnist-full}}, but without the flexible and parametric clustering offered by the copulas which is of interest for risk stratification in e.g. clinical applications. Our paper primarily focuses on the thorough evaluation of the benefits of our more flexible parametric clustering model. The adaptation of our work to the broader framework of deep clustering is not straightforward and left for future work.

\paragraph{Acknowledgements} The authors acknowledge the support from the French ANR (LABEX PRIMES of Univ. Lyon [ANR-11-LABX-0063] within the program ``Investissements d'Avenir'' [ANR-11-IDEX-0007], the JCJC project ``MIC-MAC'' [ANR-19-CE45-0005]), and the Fédération Francaise de Cardiologie (``MI-MIX'' project, Allocation René Foudon). They are also grateful to P. Croisille and M. Viallon (CREATIS, CHU Saint Etienne, France) for providing the imaging data for the HIBISCUS-STEMI population, \blueC{to} L. Petrusca and P. Clarysse (CREATIS, Villeurbanne, France) for discussions around these imaging data\blueC{, and to R. Deleat-Besson (CREATIS, Villeurbanne, France) for his help on the miniatures in Appendix C, Figure \ref{fig:infarctDisplay}}. 

\paragraph{Declaration of generative AI and AI-assisted technologies in the writing process}

Nothing to disclose.

\paragraph{Code and data availability}

The code and a demo on synthetic data corresponding to the experiments in Section \ref{sec:Experimentation on synthetic data} will be made publicly available upon acceptance.
The experiments in Section \ref{subsec:MNIST} were based on data from the public MNIST database \cite{Lecun:1998}.
The experiments in Section \ref{subsec:Application to subgroup clustering for myocardial infarct} were based on real data from the private HIBISCUS-STEMI study (ClinicalTrials ID: NCT03070496), and cannot be directly shared. However, infarct segmentations are part of the ``M1'' and ``M12'' images of the MYOSAIQ segmentation challenge dataset\footnote{\url{https://www.creatis.insa-lyon.fr/Challenge/myosaiq/}}, whose access may be provided upon request and agreement from the challenge organizers.

\paragraph{CRediT authorship contribution statement}

\begin{itemize}
    \item Fei Zheng: Methodology, Software, Validation, Investigation, Visualization, Writing - Original Draft.
    \item Nicolas Duchateau: Conceptualization, Methodology, Validation, Supervision, Writing - Review \& Editing.
\end{itemize}



\newpage

\renewcommand\thefigure{A\arabic{figure}}    
\setcounter{figure}{0}  
\renewcommand\thetable{A\arabic{table}}    
\setcounter{table}{0}  

\section*{Appendix A: \blueCN{Marginals} and copulas used in this study}
\label{sec:Margins and copulas used in this study}

We format the parameters of the studied marginal distributions by four parameters $\left\{loc, scale, \blueC{\theta_1}, \blueC{\theta_2} \right\}$, where ``$loc$'',``$scale$'' represent the location and the scale of the distribution respectively from its standardized form (with $loc=0$, $scale=1$). In detail, $loc$ and $scale$ shift and scale a variable from its standardized form by $y = (y-loc)/scale$, and its pdf $f = f/scale$. ``$\blueC{\theta_1}$'' and ``$\blueC{\theta_2}$'' are other parameters which define the shape of the distribution, one or even two of them can be absent often when the distribution is symmetric, such as Gaussian and Fisk. The studied marginal distributions are listed in Table~\ref{tab:Marginal distributions studied in this article}. The studied copulas are all single-parameter bivariate copulas, listed in Table~\ref{tab:Copulas studied in this article}. \blueC{We focused on commonly used distribution forms, varied enough and not too complex to implement. Naturally, our models remains generic and applicable to other marginal distributions and copulas.}

\renewcommand\cellset{\renewcommand\arraystretch{0.6}%
    \setlength\extrarowheight{0pt}}
\begin{table}[htbp]
	\centering
    \caption{Standardized marginal distributions studied in this article.}
    \label{tab:Marginal distributions studied in this article}
    \resizebox{\columnwidth}{!}{
    \begin{tabular}{c||c |c| c}
    \hline
    Form & PDF $f\seq{y}$ & CDF $F\seq{y}$ & parameters \\
    \hline
    Gaussian\tablefootnote{$\errorfun{x}= \frac{2}{\sqrt{\pi}}\int_0^x \exp{-t^2}dt$ represents the error function.} & $f = \frac{1}{\sqrt{2\pi}} \exp{-\frac{y^2}{2}} $ & $F = \frac{1}{2}\left(1+ \errorfun{\frac{y}{\sqrt{2}}}\right) $ & - \\
    \hline
    Gamma\tablefootnote{$\Gammafun{x} = \int_0^\infty t^{x-1}\exp{-t}dt$ is a complete Gamma function and $\incGammafun{s}{x}=\int_0^x t^{s-1}\exp{-t}dt$ represents the lower incomplete gamma function.} & $f = \frac{ y^{\blueC{\theta_1}-1} \exp{-y} }{\Gammafun{\blueC{\theta_1}}}$ & $F = \frac{\incGammafun{\blueC{\theta_1}}{y}}{\Gammafun{\blueC{\theta_1}}}$ & $\blueC{\theta_1} > 0$ \\
    \hline
    Beta\tablefootnote{$\Betafun{x}{s}=\int_0^1 t^{x-1}\left(1-t\right)^{s-1} dt$ is the Beta function and $\incBetafun{x}{a}{b}=\int_0^x t^{a-1} \left( 1-t \right)^{b-1}dt$ is the incomplete Beta function.} & $f = \frac{\Gammafun{\blueC{\theta_1}+\blueC{\theta_2}} y^{\blueC{\theta_1}-1} \left(1-y\right)^{\blueC{\theta_2}-1}   }{\Gammafun{\blueC{\theta_1}}\Gammafun{\blueC{\theta_2}} }$ & $F = \frac{\incBetafun{y}{\blueC{\theta_1}}{\blueC{\theta_2}}}{\Betafun{\blueC{\theta_1}}{\blueC{\theta_2}}}$ & \makecell{$\blueC{\theta_1}>0$, \\$\blueC{\theta_2}>0$}  \\
    \hline
    Beta prime & $f = \frac{y^{\blueC{\theta_1}-1} \left(1+y\right)^{-\blueC{\theta_1}-\blueC{\theta_2}} }{ \Betafun{\blueC{\theta_1}}{\blueC{\theta_2}} }$ & $F = \incBetafun{\frac{y}{1+y}}{\blueC{\theta_1}}{\blueC{\theta_2}} $ & \makecell{$\blueC{\theta_1}>0$, \\$\blueC{\theta_2}>0$} \\ 
    \hline
    Fisk & $f = \frac{\blueC{\theta_1} y^{\blueC{\theta_1}-1}}{\left( 1+y^{\blueC{\theta_1}} \right)^2} $  & $F = \frac{1}{1+y^{-\blueC{\theta_1}}}$ & $\blueC{\theta_1} >0$ \\
    \hline
    Laplace & $f = \frac{1}{2}\exp{ -|x| }$ & $F = \left\{ \begin{array}{ll} \frac{1}{2}\exp{y}& if \ y<0 \\ 1-\frac{1}{2}\exp{-y} & if \ y\geq 0  \end{array} \right.$ & - \\
    \hline
    Student’s t & $f=\frac{\Gammafun{\frac{\seq{\blueC{\theta_1}+1}}{2}}}{\sqrt{\pi \blueC{\theta_1}}\Gammafun{\frac{\blueC{\theta_1}}{2}}} {\seq{1+\frac{y^2}{\blueC{\theta_1}}}}^{-\frac{\blueC{\theta_1}+1}{2}}$ & $F= \frac{1}{2}+y\Gammafun{\frac{\seq{\blueC{\theta_1}+1}}{2}}\times  \frac{\hypergeofun{\frac{1}{2}}{\frac{\blueC{\theta_1}+1}{2}}{\frac{3}{2}}{-\frac{y^2}{\blueC{\theta_1}}}}{\sqrt{\pi \blueC{\theta_1}}\Gammafun{\frac{\blueC{\theta_1}}{2}}}$ & $\blueC{\theta_1} >0$ \\
    \hline
    \end{tabular}}
\end{table}

\begin{table}[htbp]
	\centering
    \caption{Copulas studied in this article.}
    \label{tab:Copulas studied in this article}
    \resizebox{\textwidth}{!}{
    \begin{tabular}{c||c c c}
    \hline
    Name & PDF $c\seq{u_1, u_2}$ & CDF $C\seq{u_1, u_2}$ & parameter \\
    \hline
    \blueC{Gaussian}\tablefootnote{$\xi_i=\phi^{-1}\seq{u_i}$, where $\phi$ represents the standard normal distribution. Correlation $\rho=\begin{bmatrix}1 & \alpha \\ \alpha & 1 \end{bmatrix}$ and $\bs{I}$ is the identity matrix.} & $c = \frac{1}{1-\alpha^2}\exp{\blueC{-\frac{1}{2}} \xi^\intercal(\rho-\bs{I})\xi }$ & $C = \int_0^{u_1} \phi \seq{ \frac{\phi^{-1}(u_2)-\rho\phi^{-1}(u) }{\sqrt{1-\rho^2}}du }$  & $\alpha\in\interv{-1,1}$ \\
    \hline
    Gumbel\tablefootnote{$U_1=\seq{- \Ln \seq{u_1}}^{\alpha}$ and $U_2=\seq{-\Ln\seq{u_2}}^{\alpha}$.} & \makecell{$ c=\frac{U_1}{u_1 \Ln(u_1)}\frac{U_2}{u_2 \Ln(u_2)} \seq{ \alpha-1+U_1+U_2}^{\frac{1}{\alpha}}$\\ \quad $ \seq{U_1+U_2}^{\frac{1}{\alpha}-2} \exp{ -\seq{U_1+U_2}^{\frac{1}{\alpha}}} $} & $C = \exp{- \seq{U_1+U_2 }^{\frac{1}{\alpha}} }$  & $\alpha\in \interv{1,+\infty}$ \\
    \hline
    Clayton & \makecell{$c = (1+\alpha)u_1^{-1-\alpha}u_2^{-1-\alpha}$ \\ \quad $\seq{ -1+u_1^{-\alpha}+u_2^{-\alpha}}^{-\frac{1}{\alpha}-2}$}  & $C = \seq{ u_1^{-\alpha}+u_2^{-\alpha}-1 }^{\frac{1}{\alpha}}$  & $\alpha \in\left[0,+\infty\right)$ \\
    \hline
    FGM & $c = 1+\alpha\seq{ 1-2u_1 }\seq{ 1-2u_2 }$ & $C = u_1 u_2 \seq{ 1+\alpha\seq{1-u_1}\seq{1-u_2}}$ & $\alpha\in \interv{-1,1}$ \\
    \hline
    Arch12\tablefootnote{$U_1 = \seq{ \frac{1}{u_1}-1 }^{\alpha}$ and $U_2 = \seq{\frac{1}{u_2}-1 }^{\alpha}$.} & \makecell{$c = \frac{U_1}{u_1(u_1-1)}\frac{U_2}{u_2(u_2-1)}\frac{ \seq{U_1+U_2}^{\frac{1}{\alpha}-2} }{ \seq{1+ \seq{U_1+U_2}^{\frac{1}{\alpha}}}^{3} }$ \\ \quad $\seq{\alpha-1+(\alpha+1)\seq{U_1+U_2}^{\frac{1}{\alpha}} }$}& $C =\seq{1+ \seq{ U_1+U_2}^{\frac{1}{\alpha}} }^{-1}$  & $\alpha\in\left[1,+\infty\right)$\\
    \hline
    Arch14\tablefootnote{$U_1=\seq{ u_1^{-\frac{1}{\alpha}} }^{\alpha}$ and $U_2=\seq{ u_2^{-\frac{1}{\alpha}} }^{\alpha}$.} & \makecell{ $c = U_1 U_2\seq{ 1+\seq{U_1+U_2}^{\frac{1}{\alpha}} }^{-2-\alpha}$ \\ \quad\qquad $\seq{U_1+U_2}^{\frac{1}{\alpha}-2}  \frac{\alpha-1+2\alpha\seq{U_1+U_2}^{\frac{1}{\alpha}}}{\alpha u_1 u_2 \seq{ u_1^{\frac{1}{\alpha}}-1 }\seq{ u_2^{\frac{1}{\alpha}}-1 }} $ }& $C =\seq{1+ \seq{ U_1+U_2}^{\frac{1}{\alpha}} }^{-\alpha}$ & $\alpha\in\left[1,+\infty\right)$ \\
    \hline
    Product & $c = 1$  & $C = u_1 u_2$ & - \\
    \hline
    \end{tabular}}
\end{table}

~\\

\newpage

\renewcommand\thefigure{B\arabic{figure}}    
\setcounter{figure}{0}  
\renewcommand\thetable{B\arabic{table}}    
\setcounter{table}{0}  

\section*{Appendix B: Convergence of the goodness of fit and error ratio when testing GICE}
\label{app:convergence}

\begin{figure}[!h]
\centering
\begin{subfigure}{0.49\textwidth}
  \centering
  \includegraphics[width=0.95\textwidth]{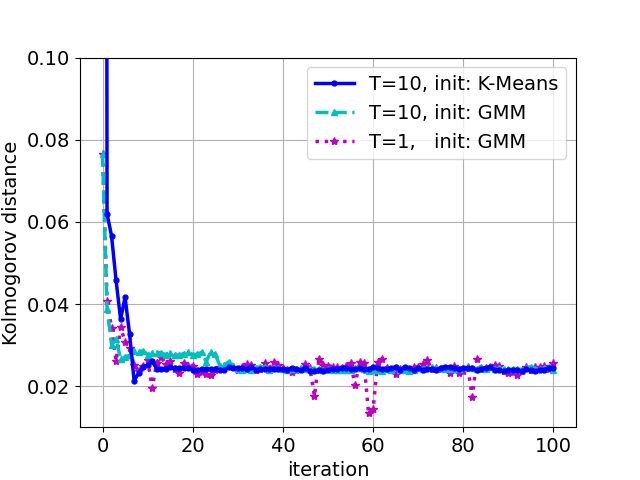}
  \captionsetup{width=.9\linewidth}
  \caption{Evolution of Kolmogorov distance}
  \label{fig:nonGauss gof}
\end{subfigure}%
\begin{subfigure}{0.49\textwidth}
  \centering
  \includegraphics[width=0.95\textwidth]{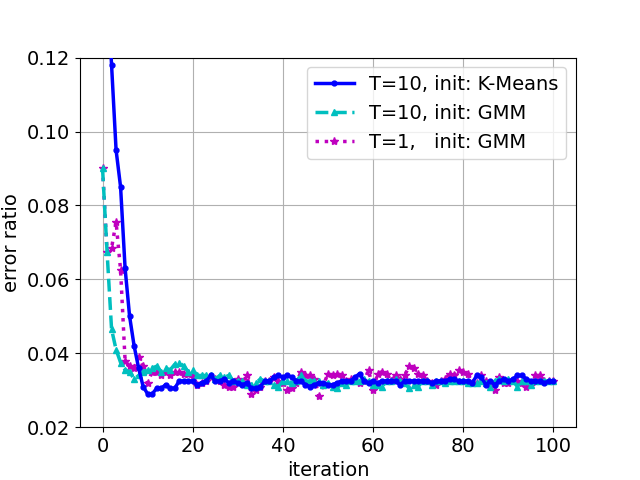}
  \captionsetup{width=.95\linewidth}
  \caption{Evolution of error ratio.}
  \label{fig:nonGauss error ratio}
\end{subfigure}
\caption{Convergence of the goodness of fit and error ratio when testing GICE on non-Gaussian CBMM synthetic data \blueCN{(N=2000 samples)}. \blueC{$T$: realization time, init: initialization method}.}
\label{fig:nonGauss CBMM convergence}
\end{figure}

\begin{figure}[!h]
\centering
\begin{subfigure}[t]{0.49\textwidth}
  \centering
  \includegraphics[width=0.95\textwidth]{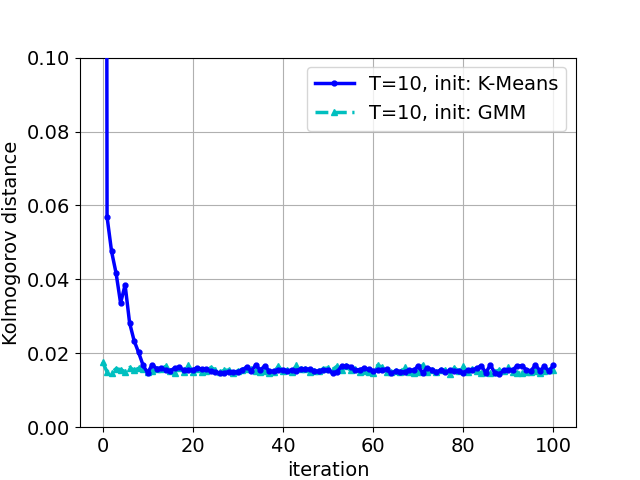}
  \captionsetup{width=.9\linewidth}
  \caption{Evolution of Kolmogorov distance}
  \label{fig:GMM gof}
\end{subfigure}%
\begin{subfigure}[t]{0.49\textwidth}
  \centering
  \includegraphics[width=0.95\textwidth]{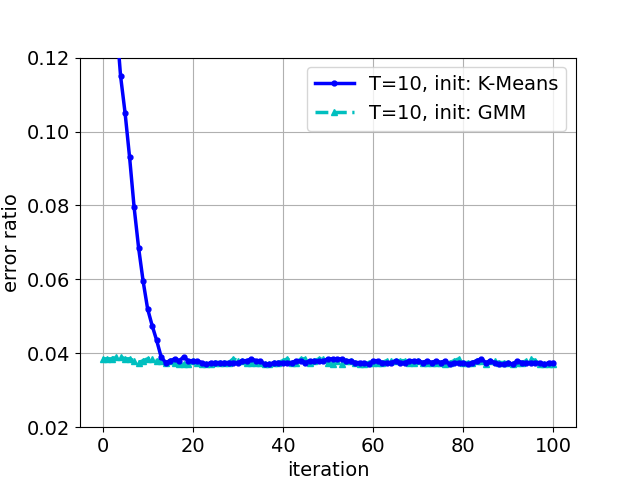}
  \captionsetup{width=.95\linewidth}
  \caption{Evolution of error ratio.}
  \label{fig:GMM error ratio}
\end{subfigure}
\caption{Convergence of the goodness of fit and error ratio when testing GICE on GMM synthetic data \blueCN{(N=2000 samples)}. \blueC{$T$: realization time, init: initialization method}.}
\label{fig:GMM convergence}
\end{figure}

~\\

\newpage

\renewcommand\thefigure{C\arabic{figure}}    
\setcounter{figure}{0}  
\renewcommand\thetable{C\arabic{table}}    
\setcounter{table}{0}  

\section*{Appendix C: Details on the imaging data and processing for characterization of myocardial patterns}
\label{app:infarct}

\begin{figure}[!h]
    \centering    
    \includegraphics[width=.9\textwidth]{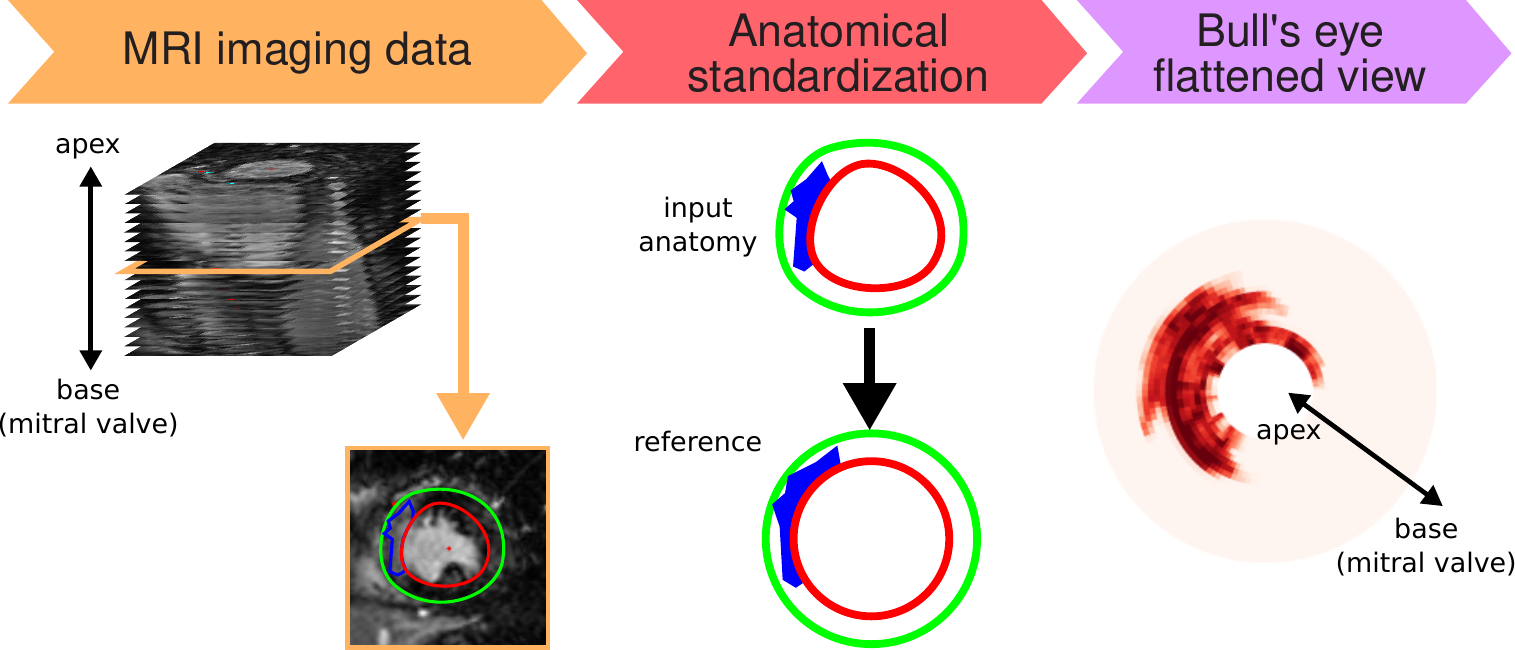}
    \caption{\blueC{Overview of the processing steps to analyze and visualize infarct patterns. The myocardium (red and green contours) and the infarct (blue area) were segmented on each slice of late Gadolinium enhancement magnetic resonance images (left). Then, the segmented data for each individual were transported to a common reference anatomy (middle). To easily visualize these 3D data, we used a Bull's eye flattened representation close to the one commonly used by clinicians (right), the intensity at each location standing for the amount of infarct transmurality.}}
    \label{fig:infarctDisplay}
\end{figure}

\begin{figure}[!h]
\centering
\begin{subfigure}[t]{0.32\textwidth}
  \centering
  \includegraphics[width=0.81\textwidth]{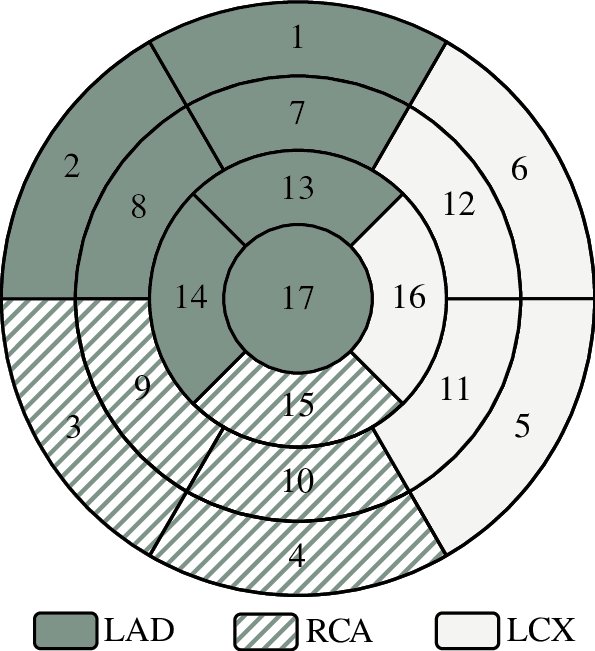}
  \captionsetup{width=.9\linewidth}
  \caption{}
  \label{fig:territory}
\end{subfigure}%
\begin{subfigure}[t]{0.32\textwidth}
  \centering
  \includegraphics[width=0.95\textwidth]{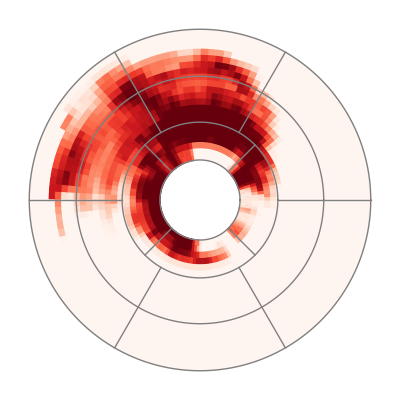}
  \captionsetup{width=.9\linewidth}
  \caption{}
  \label{fig:LAD example}
\end{subfigure}
\begin{subfigure}[t]{0.32\textwidth}
  \centering
  \includegraphics[width=0.95\textwidth]{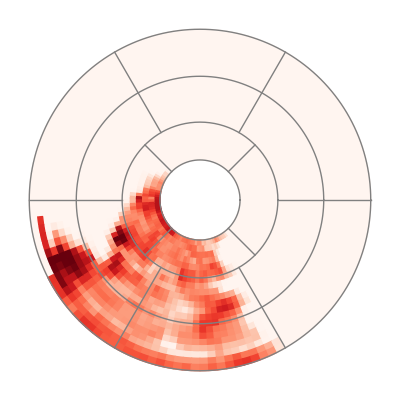}
  \captionsetup{width=.9\linewidth}
  \caption{}
  \label{fig:RCA example}
\end{subfigure}
\caption{\blueC{(a) Bull's eye flattened representation of the left ventricle, with the standard 17 segments commonly used by clinicians, which correspond to the coronary territories highlighted. (b) Bull's eye flattened representation of one representative subject from our study with LAD infarct. The 17 segments are overlaid for comprehension purposes, but our Bull's eye representation is much more detailed up to the voxel scale. (c) Similar representation for a \blueCN{subject with RCA infarct}.}}
\label{fig:Left ventricular territories and two example}
\end{figure}

\begin{figure}[!h]
    \centering
    \centering
    \includegraphics[width=\textwidth]{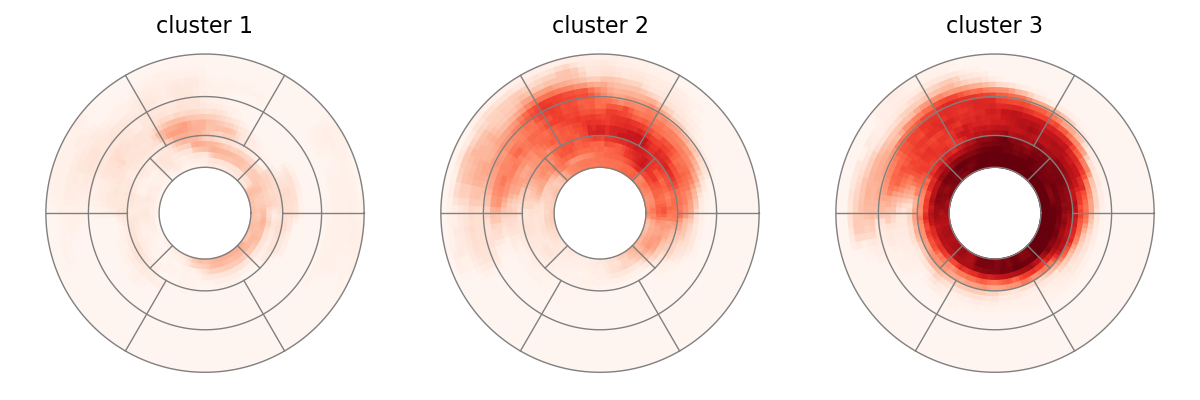}
    \label{fig:typical patterns GMM}
    \caption{Representative infarct patterns reconstructed from the ``centers'' of each \blueC{estimated CBMM }cluster. Reconstruction was carried out by multiscale kernel regression \cite{duchateau2013adaptation}.}
    \label{fig:typical patterns}
\end{figure}

\end{document}